\pdfoutput=1
\documentclass[10pt,twocolumn,letterpaper]{article}

\usepackage{preprint}
\usepackage{microtype}
\usepackage{array}
\usepackage{placeins}
\usepackage{tikz}
\usetikzlibrary{spy,arrows.meta}
\tikzset{every node/.append style={outer sep=0pt}}%

\newcommand{\sci}[2]{\ensuremath{#1\mathrm{e}{#2}}}
\renewcommand{\paragraph}[1]{\vspace{.5em}\noindent\textbf{#1.}}
\def\modelname{AnyCam4D\xspace}
\makeatletter\let\inputrows\@@input\makeatother  %
\newlength{\dynw}

\newcommand{\projecturl}{https://sascha-kirch.github.io/anycam4d/}

\usepackage{xr}
\definecolor{linkblue}{rgb}{0.21,0.49,0.74}
\usepackage[pagebackref,breaklinks,colorlinks,allcolors=linkblue]{hyperref}
\hypersetup{pdftitle={AnyCam4D: Ray-Based 4D Gaussian Rendering for Arbitrary Camera Models},
  pdfauthor={Sascha Kirch, Rafael Pagés, Jan Ondřej, Clara Pérez-Molina, Sergio Martin}}
\usepackage[capitalize]{cleveref}
\crefname{section}{Sec.}{Secs.}
\Crefname{section}{Section}{Sections}
\crefname{table}{Tab.}{Tabs.}
\Crefname{table}{Table}{Tables}
\usepackage{orcidlink}

\title{AnyCam4D: Ray-Based 4D Gaussian Rendering for Arbitrary Camera Models}

\author{
Sascha Kirch\,\orcidlink{0000-0002-5578-7555}$^{1}$ \quad
Rafael Pagés\,\orcidlink{0000-0002-5691-9580}$^{2}$ \quad
Jan Ondřej\,\orcidlink{0000-0002-5409-1521}$^{2}$ \quad
Clara Pérez-Molina\,\orcidlink{0000-0001-8260-4155}$^{1}$ \quad
Sergio Martin\,\orcidlink{0000-0002-4118-0234}$^{1}$ \\[4pt]
$^{1}$UNED -- Universidad Nacional de Educación a Distancia, Madrid, Spain \\
$^{2}$Volograms, Dublin, Ireland \\
{\small Project page: \url{\projecturl}}
}

\begin{document}
\twocolumn[{%
\renewcommand\twocolumn[1][]{#1}%
\maketitle
\begin{center}
\includegraphics[width=\textwidth]{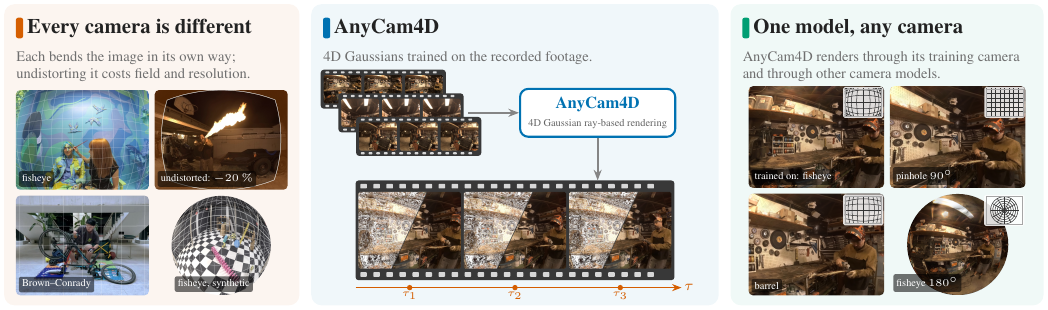}
\captionof{figure}{\emph{Every camera is different}: lenses bend the image in different ways, and undistorting the footage costs field of view and resolution (shaded: the $20\,\%$ of a fisheye frame outside its released pinhole version). \emph{AnyCam4D} trains 4D Gaussians on the footage as recorded, through each camera's calibrated pose and lens model, and renders them by ray-based evaluation; the strip shows renders and Gaussians at three time instants $\tau$. \emph{One model, any camera}: the model renders through its training camera and through other camera models.}\label{fig:teaser}
\end{center}
\vspace{0.5em}

}]
\begin{abstract}
Real cameras are not pinholes, yet most dynamic Gaussian splatting assumes they are.
Footage is therefore undistorted first, losing field of view or detail, especially with wide-angle and fisheye cameras.
Static Gaussian splatting has largely solved this with ray-based rendering, which suits any camera.
Native 4D Gaussians, which model motion in space-time, cannot use it directly: each Gaussian's orientation and size cannot be recovered reliably from its covariance.\\
We present \modelname, which trains native 4D Gaussians on original footage by ray-based rasterization, preserving field of view and resolution at about the per-step cost of standard projection.
Our key contribution is a numerically stable way to obtain orientation and size: we condition a square root of the 4D covariance.
Trained models render through other cameras, with rolling shutter, depth of field or chromatic aberration.
With a synthetic $170^\circ$ fisheye camera, training on original images improves peak signal-to-noise ratio (PSNR) by 3.0--8.1\,dB.
On real footage, \modelname has lower perceptual error than the standard projection adapted to each camera, at comparable PSNR and without camera-specific derivations.
With the synthetic camera rotated $60$--$80^\circ$ away from the capture rig's view, it leads by 6.2\,dB against ground truth.
\end{abstract}

\section{Introduction}\label{sec:intro}

Dynamic Gaussian splatting~\cite{4dgs_yang,4dgs_wu,stg,freetimegs} reconstructs moving scenes from multi-camera video and renders new views photorealistically in real time, but usually assumes pinhole cameras.
Its projection, EWA splatting~\cite{ewa}, linearizes the camera at each Gaussian's mean and orders Gaussians by depth along the optical axis, which is accurate only near the axis of an undistorted camera.
Real lenses distort, from the mild radial distortion of standard lenses to fisheye lenses that see $180^\circ$ and more, and rigs, for example in robotics, often combine several.

Footage is therefore undistorted first, at the cost of field or resolution.
At equal resolution a pinhole image either crops the field of a wide lens or samples its center more coarsely: undistorting a $170^\circ$ fisheye to keep $150^\circ$ across the image samples its center 3.4 times more coarsely per axis, keeping the center's resolution leaves $95^\circ$, and beyond $180^\circ$ no pinhole image exists.

For static scenes, ray-based renderers~\cite{3dgrt,3dgut,3dgeer} have removed this limitation: they evaluate each Gaussian along the ray of a pixel, so the camera only has to provide rays, and 3DGUT~\cite{3dgut} keeps rasterization speed by selecting tiles with an unscented transform (UT).
Dynamic scenes cannot simply swap in such a renderer: the 3D Gaussian that a native 4D Gaussian~\cite{4dgs_yang} draws at time $\tau$ comes from conditioning, which yields a covariance but not the rotation and scale that the UT and the ray evaluation are built from.
Recovering them is unreliable: an eigendecomposition has no usable gradient at the usual isotropic initialization, at any precision, and in the single precision (fp32) of GPU rasterizers the conditioned covariance loses the thin axis of fast, thin Gaussians.

We present \modelname, which trains and renders native 4D Gaussians through any camera model.
Conditioning the square root of the 4D covariance, rather than the covariance itself, keeps the thin axis and yields a triangular factor that replaces rotation and scale throughout.
Since a camera only has to provide a projection and a ray per pixel, one model also renders through other cameras, with rolling shutter, depth of field or chromatic aberration.
On a synthetic $170^\circ$ fisheye, training on footage as recorded gains 3.0--8.1\,dB in peak signal-to-noise ratio (PSNR), depending on the field the undistortion keeps.
Against EWA given the native lens's Jacobian, \modelname has comparable PSNR and lower perceptual error (LPIPS) on real footage, at about the same cost per training step; on the synthetic fisheye, with a held-out camera rotated by $60$--$80^\circ$, it leads by 6.2\,dB.

\noindent In summary, our contributions are:
\begin{itemize}
\item \modelname, a ray-based rasterizer that trains native 4D Gaussians through any camera model at about EWA's cost per step and memory per Gaussian, and renders them through other cameras.
\item Factored conditioning: conditioning the square root of the 4D covariance gives an exact triangular factor that, with a ray-restricted, closest-point evaluation, keeps ray evaluation and its gradients accurate in fp32.
\item A controlled evaluation that separates native training from projection, with a native-lens EWA baseline, a rotation probe and path-traced ground truth on a synthetic scene.
\end{itemize}

\section{Related Work}\label{sec:related}

\begin{table}[t]
\centering
\caption{Scope of related methods. \emph{Dyn.}: dynamic scenes. \emph{4D}: the primitive is a 4D Gaussian with a full space-time covariance. \emph{Cam.}: arbitrary camera model. \emph{Dist.}: trains on distorted footage. \emph{RS}, \emph{DoF}: rolling shutter and depth of field. ($\checkmark$): render only. warp: a pinhole render warped into the distorted image. $^*$Same primitive in other parameters (\cref*{S-sec:s_freetime}).}\label{tab:t1}
\small
\setlength{\tabcolsep}{3.5pt}
\begin{tabular}{@{}lcccccc@{}}
\toprule
Method & Dyn. & 4D & Cam. & Dist. & RS & DoF \\
\midrule
4DGS~\cite{4dgs_yang} & $\checkmark$ & $\checkmark$ & & & & \\
4DGaussians~\cite{4dgs_wu} & $\checkmark$ & & & & & \\
4D-Rotor~\cite{4d_rotor} & $\checkmark$ & $\checkmark$ & & & & \\
STGS~\cite{stg} & $\checkmark$ & & & warp & & \\
FreeTimeGS~\cite{freetimegs} & $\checkmark$ & $\checkmark^*$ & & & & \\
Fisheye-GS~\cite{fisheye_gs} & & & fisheye & $\checkmark$ & & \\
DirectFisheye-GS~\cite{directfisheye_gs} & & & fisheye & $\checkmark$ & & \\
3DGRT~\cite{3dgrt} & & & $\checkmark$ & $\checkmark$ & $\checkmark$ & ($\checkmark$) \\
3DGUT~\cite{3dgut} & & & $\checkmark$ & $\checkmark$ & $\checkmark$ & \\
3DGEER~\cite{3dgeer} & & & $\checkmark$ & $\checkmark$ & & \\
SplatAD~\cite{splatad} & rigid & & & & $\checkmark$ & \\
4D-GRT~\cite{4dgrt} & $\checkmark$ & & ($\checkmark$) & & ($\checkmark$) & ($\checkmark$) \\
\midrule
\modelname{} (ours) & $\checkmark$ & $\checkmark$ & $\checkmark$ & $\checkmark$ & ($\checkmark$) & ($\checkmark$) \\
\bottomrule
\end{tabular}
\end{table}

Camera-general Gaussian splatting has so far addressed static scenes, and dynamic methods render through pinholes, at most warped into the distorted image afterwards; \modelname combines both.

\subsection{Camera Models in Gaussian Splatting}

3D Gaussian splatting~\cite{3dgs} reached real-time rendering by projecting each Gaussian with EWA splatting~\cite{ewa}, a local affine approximation of the camera, and sorting along the optical axis.
Both steps assume a pinhole, so each further lens needs its own Jacobian.
Fisheye-GS~\cite{fisheye_gs} derives it for the equidistant model, DirectFisheye-GS~\cite{directfisheye_gs} trains on raw fisheye images and a self-calibrating approach~\cite{selfcal_gs} learns the lens.
Ray-based renderers drop the projection Jacobian altogether: 3DGRT~\cite{3dgrt} traces rays through a bounding-volume hierarchy, 3DGUT~\cite{3dgut} rasterizes with tiles selected by a UT, and 3DGEER~\cite{3dgeer} integrates each Gaussian exactly along the ray.
Gunes \etal~\cite{fisheye_fov_eval} compare Fisheye-GS and 3DGUT on real fisheye captures up to $200^\circ$.

\subsection{Dynamic Gaussian Scenes}

Dynamic Gaussian methods, which succeed radiance fields with deformation fields~\cite{dnerf} or space-time planes~\cite{hexplane,kplanes}, fall into two families.
Deformation methods warp a canonical set of 3D Gaussians per frame, with an MLP~\cite{deformable_3dgs}, a HexPlane~\cite{4dgs_wu}, or sparse control points~\cite{scgs}.
Native methods put time into the primitive: 4DGS~\cite{4dgs_yang,4dgs_native} and 4D-Rotor~\cite{4d_rotor} carry a full 4D covariance and condition it on the query time, while Spacetime Gaussians (STGS)~\cite{stg}, FreeTimeGS~\cite{freetimegs}, Ex4DGS~\cite{ex4dgs} and RetimeGS~\cite{retimegs} attach temporal opacity and motion to each Gaussian, and SaRO-GS~\cite{saro_gs} adds a residual field.
The native 4D Gaussian remains a common base, for long sequences~\cite{longvolcap}, faster rendering~\cite{4dgs_1k} and faster training~\cite{instant4d}.
These methods keep the EWA projection onto a pinhole image; STGS trains on distorted fisheye footage by warping its pinhole render into the fisheye image~\cite{stg}. The camera-general renderers above have not been carried over to them, and the efficiency and densification advances of static splatting~\cite{taming3dgs,absgs,revising_densification,pixelgs,mcmc} only in part.

\subsection{Ray-Based Dynamic Rendering}

4D-GRT~\cite{4dgrt} is closest to our work: it combines the deformation field of 4DGaussians~\cite{4dgs_wu} with the ray tracer of 3DGRT and renders fisheye distortion, depth of field and rolling shutter from the trained model.
SplatAD~\cite{splatad} trains through rolling shutter for driving scenes, with rigid actors moved per frame rather than a deforming scene.
4D-GRT differs from \modelname in three respects: (i) it deforms 3D Gaussians rather than conditioning native 4D ones, (ii) it trains on pinhole images only, and (iii) it is not compared against EWA on the same data.
\Cref{tab:t1} compares the capabilities of these methods with ours.

\section{Preliminaries}\label{sec:prelim}

\paragraph{Native 4D Gaussians}
4DGS~\cite{4dgs_yang} represents a dynamic scene by Gaussians in space-time.
A primitive has a mean $(\mu_{xyz}, \mu_t) \in \mathbb{R}^3 \times \mathbb{R}$, an opacity $\eta \in (0,1)$ and a covariance $\Sigma_{4D} = R S^2 R^\top$ with the scales $S = \operatorname{diag}(s_x, s_y, s_z, s_t)$.
The 4D rotation $R$ is the product of a left and a right isoclinic rotation, each given by a unit quaternion.
At render time $\tau$ the drawn object is the Gaussian conditioned on $t = \tau$.
With the blocks
\begin{equation}\label{eq:blocks}
  \Sigma_{4D} = \begin{pmatrix} \Sigma_{xyz} & \Sigma_{xyz,t} \\ \Sigma_{xyz,t}^\top & \sigma_t^2 \end{pmatrix},
\end{equation}
the conditional 3D Gaussian and the temporal gate are
\begin{equation}\label{eq:cond}
\begin{aligned}
  \mu_{3D}(\tau) &= \mu_{xyz} + \Sigma_{xyz,t}\,(\tau - \mu_t) / \sigma_t^2, \\
  \Sigma_{3D} &= \Sigma_{xyz} - \Sigma_{xyz,t}\Sigma_{xyz,t}^\top / \sigma_t^2, \\
  w_t &= \exp\bigl(-(\tau - \mu_t)^2 / 2\sigma_t^2\bigr).
\end{aligned}
\end{equation}
The gate $w_t$ multiplies the opacity.
The conditional mean moves with the constant velocity $v = \Sigma_{xyz,t}/\sigma_t^2$; the conditional covariance $\Sigma_{3D}$ does not depend on $\tau$.
Color is a 4D spherindrical harmonic: each spatial harmonic has a temporal Fourier series, so view dependence changes with time.

\paragraph{EWA splatting}
3DGS~\cite{3dgs} and 4DGS project $\Sigma_{3D}$ with the local affine approximation of EWA splatting~\cite{ewa,ewa_splatting}, $\Sigma_{2D} = J W \Sigma_{3D} W^\top J^\top$, where $W$ is the world-to-camera rotation and $J$ the Jacobian of the camera projection at the mean.
Its error grows with the extent of the Gaussian, the field angle and the nonlinearity of the lens, and each lens model needs its own Jacobian~\cite{fisheye_gs}.

\paragraph{Unscented projection and ray evaluation}
3DGUT~\cite{3dgut} replaces the linearization with a UT~\cite{unscented}.
For a 3D Gaussian with rotation $R_{3D}$ and scales $S_{3D}$, seven sigma points
\begin{equation}\label{eq:sigma}
  \mathcal{X}_0 = \mu_{3D}, \qquad \mathcal{X}_{\pm i} = \mu_{3D} \pm \sqrt{3+\lambda}\,[R_{3D}S_{3D}]_i,
\end{equation}
with $[\cdot]_i$ the $i$-th column and $\lambda$ the UT spread parameter, are mapped through an arbitrary camera function, and their weighted moments give a 2D mean and covariance.
These only select the tiles a Gaussian is tested in and its sort key.
The contribution to a pixel is evaluated in 3D, at the maximum density along the line of the pixel ray $o + \ell\,d$, with origin $o$ and unit direction $d$ from unprojecting the pixel through the camera, giving the squared Mahalanobis distance $D^2$ and the opacity $\alpha$ of the Gaussian at that pixel:
\begin{equation}\label{eq:d2}
\begin{aligned}
  D^2 &= \min_{\ell \in \mathbb{R}}\,(p + \ell d)^\top \Sigma_{3D}^{-1} (p + \ell d), \quad p = o - \mu_{3D}, \\
  \alpha &= \eta\, w_t\, e^{-D^2/2}.
\end{aligned}
\end{equation}
Since $\Sigma_{3D} = R_{3D}S_{3D}^2R_{3D}^\top$, 3DGUT evaluates $D^2$ by whitening the ray with $S_{3D}^{-1}R_{3D}^\top$, from the rotation and scales it stores as parameters.

\section{Method}\label{sec:method}

\begin{figure*}[t]
\centering
\includegraphics[width=\textwidth]{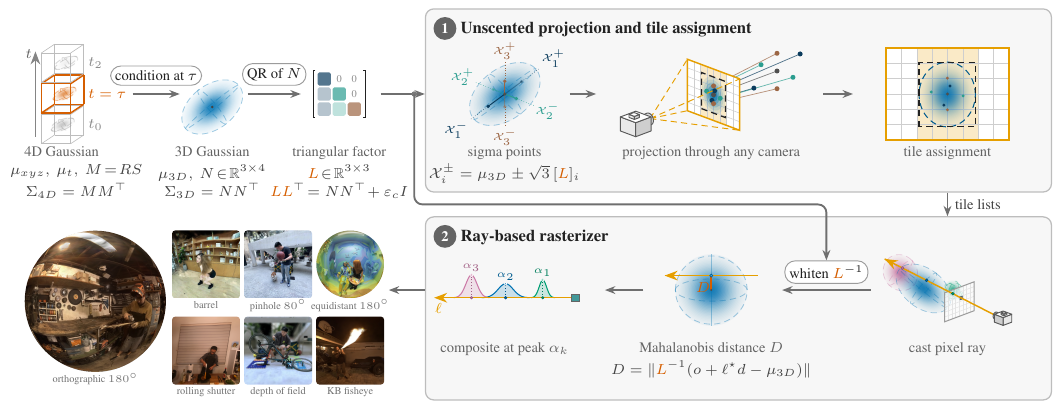}
\caption{Rendering pipeline. The square root $M$ of the 4D covariance is conditioned on $\tau$, giving the square root $N$ of the 3D covariance (\cref{eq:factored}); a QR decomposition turns $N$ into the triangular factor $L$, with $LL^\top = NN^\top + \varepsilon_c I$. (1) Sigma points are projected through any camera to assign tiles. (2) Each pixel ray is whitened by $L^{-1}$ and Gaussians are composited at their peak opacity along it.}\label{fig:f2}
\end{figure*}

\modelname evaluates each native 4D Gaussian along the rays of any camera model that maps points to pixels and back; a UT inspired by 3DGUT only selects its tiles.
Both steps need the rotation and scales of a 3D Gaussian, which a conditioned 4D Gaussian does not have.

\subsection{Ray Evaluation of Native 4D Gaussians}\label{sec:lift}

\paragraph{Why the naive route fails}
The two steps of 3DGUT use $\Sigma_{3D}$ in different forms.
The sigma points (\cref{eq:sigma}) step along the columns of a square root $B$ with $BB^\top = \Sigma_{3D}$, and the distance $D^2$ (\cref{eq:d2}) applies $\Sigma_{3D}^{-1} = B^{-\top}B^{-1}$ to the ray.
A 3D Gaussian has both at hand, since its stored rotation and scales give $B = R_{3D}S_{3D}$ and $B^{-1} = S_{3D}^{-1}R_{3D}^\top$.
A conditioned 4D Gaussian stores neither: conditioning (\cref{eq:cond}) yields only the covariance $\Sigma_{3D}$.

The direct way to recover rotation and scales is an eigendecomposition $\Sigma_{3D} = V\Lambda V^\top$, with $R_{3D} = V$ and $S_{3D} = \sqrt{\Lambda}$.
Its gradient contains terms $1/(\lambda_i - \lambda_j)$, which diverge for equal axes, as in the usual isotropic initialization.

The covariance itself is already unreliable for a fast, thin Gaussian, long in space-time but thin at any one time: computing $\Sigma_{3D}$ by \cref{eq:cond} subtracts two nearly equal matrices.
With the seven significant digits of fp32, the thin axis is then lost to rounding and its variance can even come out negative, so that $\Sigma_{3D}$ is no longer a valid covariance; condition numbers reach \sci{7}{8} in our checkpoints.
The same loss reaches every other factor of this $\Sigma_{3D}$, including its inverse and its Cholesky factor (\cref{tab:routes}).

\paragraph{Conditioning the square root}
Unlike the conditioned 3D Gaussian, the 4D Gaussian stores rotation and scales, so a square root of its covariance is available: $M = RS$ with $\Sigma_{4D} = MM^\top$.
We therefore write $\Sigma_{3D}$ of \cref{eq:cond} in terms of $M$, never computing that subtraction:
\begin{equation}\label{eq:factored}
\begin{aligned}
  M &= \begin{pmatrix} M_{xyz} \\ m^\top \end{pmatrix}, \\
  N &= M_{xyz}\bigl(I - \hat m \hat m^\top\bigr), \qquad \Sigma_{3D} = N N^\top,
\end{aligned}
\end{equation}
with $M_{xyz} \in \mathbb{R}^{3\times4}$, $m \in \mathbb{R}^{4}$ and the unit time direction $\hat m = m / \lVert m \rVert$.
The orthogonal projector $I - \hat m \hat m^\top$ removes the time component from every axis.
This equals $\Sigma_{3D}$ of \cref{eq:cond} exactly (\cref*{S-sec:s_factor_fwd}).
$N$ holds axis lengths instead of their squares, which is what preserves the thin axis: at \sci{1}{-4} of the longest axis it stays three digits above fp32 rounding, while its variance, at \sci{1}{-8}, falls below it.

\begin{table*}[t]
\centering
\caption{Novel-view quality on the held-out cameras at the last training step, mean over three seeds; per-scene results with seed ranges and SSIM in \cref*{S-tab:s_quality}.
\emph{Undistorted target}: trained and scored on undistorted images.
\emph{Native target}: scored on the shared target of \cref{sec:setup}; {\modelname}\,ud is drawn through the native lens, EWA\,nat projects through the Jacobian of the native lens.
PSNR differences: $\Delta$ = {\modelname}\,ud $-$ EWA\,ud (renderer, undistorted), $\Delta_\text{nat}$ = {\modelname}\,nat $-$ {\modelname}\,ud (native training), $\Delta_\text{proj}$ = {\modelname}\,nat $-$ EWA\,nat (renderer).
Field: largest incidence angle on a held-out pixel of the native lens (range over the six SelfCap clips).}\label{tab:t3}
\setlength{\tabcolsep}{2.3pt}
\footnotesize
\begin{tabular}{@{}llc cc cc c cc cc cc cc@{}}
\toprule
& & & \multicolumn{5}{c}{Undistorted target} & \multicolumn{8}{c}{Native target} \\
\cmidrule(lr){4-8}\cmidrule(l){9-16}
& & & \multicolumn{2}{c}{EWA\,ud} & \multicolumn{2}{c}{\modelname\,ud} & & \multicolumn{2}{c}{\modelname\,ud} & \multicolumn{2}{c}{EWA\,nat} & \multicolumn{2}{c}{\modelname\,nat} & & \\
\cmidrule(lr){4-5}\cmidrule(lr){6-7}\cmidrule(lr){9-10}\cmidrule(lr){11-12}\cmidrule(lr){13-14}
Dataset & Lens & Field\,[$^\circ$] & PSNR$\uparrow$ & LPIPS$\downarrow$ & PSNR$\uparrow$ & LPIPS$\downarrow$ & $\Delta$ & PSNR$\uparrow$ & LPIPS$\downarrow$ & PSNR$\uparrow$ & LPIPS$\downarrow$ & PSNR$\uparrow$ & LPIPS$\downarrow$ & $\Delta_\text{nat}$ & $\Delta_\text{proj}$ \\
\midrule
DyNeRF & BC & 49 & 34.05 & .070 & 34.03 & .069 & $-$0.03 & 33.64 & .083 & 33.37 & .071 & 34.16 & .068 & $+$0.52 & $+$0.79 \\
SelfCap & BC & 36--42 & 26.65 & .166 & 26.76 & .162 & $+$0.11 & 26.28 & .190 & 25.99 & .187 & 26.12 & .179 & $-$0.15 & $+$0.14 \\
Dynopsis & BC & 54 & 29.72 & .034 & 29.90 & .032 & $+$0.18 & 28.63 & .052 & 29.17 & .031 & 29.46 & .028 & $+$0.82 & $+$0.29 \\
\midrule
DeepView & KB & 75 & 27.97 & .102 & 28.02 & .098 & $+$0.05 & 27.24 & .098 & 28.06 & .087 & 27.88 & .083 & $+$0.64 & $-$0.17 \\
Dynopsis & KB & 85 & 26.85 & .065 & 32.60 & .030 & $+$5.75 & 25.06 & .046 & 32.43 & .015 & 33.11 & .012 & $+$8.05 & $+$0.68 \\
\bottomrule
\end{tabular}
\end{table*}

\paragraph{A square factor}
$N$ is a square root, but with three rows and four columns it has no inverse, so it cannot whiten a ray, and it holds 12 numbers.
We need a $3{\times}3$ matrix $L$ with the same $LL^\top$, triangular so that its inverse is cheap to apply.
Forming $NN^\top$ and factoring it squares the condition number of $N$ and again loses thin axes, so we work on $N$ itself.
Let
\begin{equation}\label{eq:qr}
  A = \begin{pmatrix} N^\top \\ \sqrt{\varepsilon_c}\, I \end{pmatrix} \in \mathbb{R}^{7\times3}, \qquad A = QU,
\end{equation}
with four rows from the Gaussian's axes and three from the floor $\varepsilon_c = \sci{1}{-6}$, which sets the narrowest axis the renderer draws.
The QR decomposition, with $Q^\top Q = I$ and $U$ upper triangular, gives $A^\top A = U^\top U$, so $L = U^\top$ is lower triangular with $LL^\top = \Sigma_{3D} + \varepsilon_c I$.
$L$ is the Cholesky factor of the floored covariance, computed without forming that covariance, as in square-root Kalman filters~\cite{kaminski_sqrt,bierman}.
So the optimized rotation and log-scales give $M$, $N$ is the exact conditioned root through which gradients flow, and $L$ is its compact, invertible form of six numbers.

In \cref{tab:routes}, working on $N$ removes the gradient error with either factorization, and QR also the forward error that $NN^\top$ leaves on the thinnest Gaussians (none off by over 1\,\%), at the same cost.
Re-rendered this way, trained models agree with their earlier renders to at least 72.7\,dB; the gain is in training, where non-finite gradients on native lenses fell from 4 of 49 runs to none of 72 (\cref*{S-sec:s_precision}).

\begin{table}[t]
\centering
\caption{Error in fp32, against fp64, of each way to obtain the factor of $\Sigma_{3D}$ used for rendering, starting from $\Sigma_{3D}$ or from $N$. \emph{Trained models}: number of Gaussians whose distance $D^2$ is off by more than 1\,\% or 100\,\%, out of 0.6--3.1\,M per model, range over six models. \emph{Synthetic}: p99 relative error of the rotation gradient on 20\,k fast, thin Gaussians, whose $\Sigma_{3D}$ loses the thin axis. $^*$37\,\% non-finite with two equal axes. $^\dagger$Loss on $\Sigma_{3D}^{-1}$.}\label{tab:routes}
\footnotesize
\setlength{\tabcolsep}{3pt}
\begin{tabular}{@{}llrr|r@{}}
\toprule
& & \multicolumn{2}{c|}{Trained models} & Synthetic \\
\cmidrule(lr){3-4}\cmidrule(l){5-5}
Start & Method & $>$\,1\,\% & $>$\,100\,\% & gradient \\
\midrule
$\Sigma_{3D}$ (\cref{eq:cond}) & inverse & 467--4479 & 15--213 & \sci{3}{3}$^\dagger$ \\
$\Sigma_{3D}$ (\cref{eq:cond}) & eigendecomposition & 493--5828 & 6--88 & undefined$^*$ \\
$\Sigma_{3D}$ (\cref{eq:cond}) & Cholesky & 474--4467 & 6--88 & 2.03 \\
$N$ (\cref{eq:factored}) & $NN^\top$, Cholesky & 1--414 & 0--2 & \sci{6.2}{-4} \\
$N$ (\cref{eq:factored}) & QR (ours) & 0 & 0 & \sci{6.1}{-4} \\
\bottomrule
\end{tabular}
\end{table}

\paragraph{Sigma points and rays}
$L$ replaces rotation and scales throughout.
The sigma points of \cref{eq:sigma} become $\mathcal{X}_{\pm i} = \mu_{3D} \pm \sqrt{3}\,[L]_i$ with $\lambda = 0$.
Any square root gives sigma points of the same mean and covariance, and through a fisheye up to $100^\circ$ of incidence the projected covariance is as accurate as from the principal axes (\cref*{S-sec:s_numerics}).

\Cref{eq:d2} minimizes over the whole line through the pixel; we keep only the ray in front of its origin, $\ell \ge 0$, where $L$ gives the peak distance $\ell^\star$ in closed form:
\begin{equation}\label{eq:whiten}
\begin{aligned}
  \ell^\star &= \max\Bigl(0,\, -\frac{(L^{-1}d)^\top L^{-1}p}{\lVert L^{-1}d\rVert^2}\Bigr), \\
  D^2 &= \bigl\lVert L^{-1}(p + \ell^\star d)\bigr\rVert^2 .
\end{aligned}
\end{equation}
If the line's peak lies behind the origin, the ray's density peaks at $\ell^\star = 0$.
$L^{-1}$ is never computed: since $L$ is triangular, $L^{-1}x$ is solved row by row (forward substitution), which is cheaper and more accurate than an explicit inverse.
$L$ is computed once per Gaussian in each render and reused by all its rays.
Over the whole line, a long Gaussian that reaches behind the camera is drawn with the density it has there, and on wide lenses many Gaussians straddle the camera plane; restricting to the ray avoids this and gains 1.8\,dB on undistorted Dynopsis KB (\cref*{S-tab:s_numerics}).
Gaussians whose mean lies behind the camera plane are culled; tile intersection and compositing follow 3DGUT~\cite{3dgut}, sorting by distance to the camera center, which holds at any incidence.

\paragraph{Far from the camera}
Whitening divides the offset $p$ from the center to the ray origin by the Gaussian's axes, so the fp32 error of $D^2$ grows with $\lVert p \rVert$ and with the axis ratio of a thin Gaussian.
We therefore first move the ray origin to the point of the ray closest to the center, $p \leftarrow p - (p^\top d)\,d$, which leaves $D^2$, its gradient and the clamp unchanged.
Against fp64 this lowers the error of $D^2$ by a factor of 100--2000 at axis ratios of 100--1000, and in training it removed the late collapses and non-finite runs we had observed (small samples; \cref*{S-tab:s_numerics}).

\paragraph{Gradient}
The gradient of $D^2$ needs no derivative of $\ell^\star$: either $\ell^\star$ is the closest point, where $D^2$ is flat in $\ell$, or it is held at $0$.
With $r = L^{-1}(p + \ell^\star d)$ and $w = L^{-\top} r$, one more triangular solve (back substitution),
\begin{equation}\label{eq:envelope}
  \frac{\partial D^2}{\partial \mu_{3D}} = -2\,w, \qquad \frac{\partial D^2}{\partial L} = -2\,\operatorname{tril}\bigl(w\,r^\top\bigr),
\end{equation}
where $\operatorname{tril}$ keeps the lower triangle.
From $L$ the gradient flows back through the QR and \cref{eq:factored} to the 4D rotations and scales in closed form (\cref*{S-sec:s_factor}); unlike that of an eigendecomposition, it is well defined for isotropic Gaussians, since the floor keeps every diagonal entry of $L$ at least $\sqrt{\varepsilon_c}$, and going through $N$ keeps rotation gradients accurate where differentiating the subtraction in \cref{eq:cond} does not.

\begin{table*}[t]
\centering
\caption{PSNR against published methods, per scene, under each dataset's published protocol; SSIM and LPIPS in \cref*{S-tab:s_prior}.
SelfCap uses the clips of FreeTimeGS. Published numbers are taken from FreeTimeGS~\cite{freetimegs} (DyNeRF, SelfCap) and STGS~\cite{stg} (DeepView); ours are the mean over three seeds at the last step.
$^\dagger$One model per 50-frame segment, trained through a warp of a pinhole render; ours train one model per full clip through the native lens.
$^\ddagger$On DeepView, EWA through the native lens (EWA\,nat, \cref{sec:setup}).}\label{tab:t2}
\setlength{\tabcolsep}{3.4pt}
\footnotesize
\begin{tabular}{@{}l ccc cccccc ccc@{}}
\toprule
& \multicolumn{3}{c}{DyNeRF} & \multicolumn{6}{c}{SelfCap} & \multicolumn{3}{c}{DeepView (fisheye)} \\
\cmidrule(lr){2-4}\cmidrule(lr){5-10}\cmidrule(l){11-13}
Method & spinach & steak & sear & bike1 & bike2 & corgi1 & corgi2 & dance1 & dance2 & flames & truck & welder \\
\midrule
K-Planes~\cite{kplanes} & 32.60 & 32.38 & 32.52 & -- & -- & -- & -- & -- & -- & -- & -- & -- \\
HyperReel~\cite{hyperreel} & 32.30 & 32.20 & 32.57 & -- & -- & -- & -- & -- & -- & 30.63$^\dagger$ & 27.18$^\dagger$ & 25.55$^\dagger$ \\
4DGaussians~\cite{4dgs_wu} & 32.46 & 32.51 & 32.49 & 24.30 & 25.25 & 24.15 & 28.04 & 26.82 & 26.54 & -- & -- & -- \\
STGS~\cite{stg} & 33.18 & 33.64 & 33.89 & 21.46 & 22.09 & 27.55 & 28.52 & 25.26 & 23.74 & 30.57$^\dagger$ & 27.31$^\dagger$ & 26.84$^\dagger$ \\
Ex4DGS~\cite{ex4dgs} & 33.23 & 33.91 & 33.69 & -- & -- & -- & -- & -- & -- & -- & -- & -- \\
4DGS~\cite{4dgs_yang} & 32.93 & 34.03 & 33.51 & 24.04 & 24.77 & 26.95 & 27.50 & 26.40 & 25.53 & -- & -- & -- \\
FreeTimeGS~\cite{freetimegs} & 33.80 & 34.98 & 34.06 & 24.96 & 25.67 & 28.90 & 29.96 & 27.66 & 27.85 & -- & -- & -- \\
\midrule
EWA (our impl.)$^\ddagger$ & 34.34 & 33.66 & 34.16 & 25.52 & 24.24 & 28.61 & 28.10 & 27.09 & 26.35 & 28.25 & 28.10 & 27.94 \\
\modelname{} (ours) & 34.23 & 33.85 & 34.00 & 25.79 & 24.05 & 28.75 & 28.19 & 27.44 & 26.34 & 28.45 & 28.11 & 27.08 \\
\bottomrule
\end{tabular}
\end{table*}

\subsection{Appearance, Densification and Optimization}\label{sec:training}

\paragraph{Additive temporal color}
Instead of 4D harmonics, we add a direction-independent temporal term to the spatial harmonics,
\begin{equation}\label{eq:color}
  \mathbf{c}(d, \tau) = \sum_{k} Y_k(d)\, \mathbf{a}_k + \sum_{f=1}^{F} \bigl(\mathbf{b}_f \cos \omega_f + \mathbf{b}'_f \sin \omega_f\bigr),
\end{equation}
with spatial harmonics $Y_k$, learned coefficients $\mathbf{a}_k, \mathbf{b}_f, \mathbf{b}'_f$, $\omega_f = \pi f (\tau - \mu_t)$ with $\tau - \mu_t$ the Gaussian's local time, $\tau \in [-\tfrac12, \tfrac12]$ over the clip, and $F = 4$, in the spirit of the Fourier color residuals of Gaussian-Flow~\cite{gaussian_flow}; Instant4D~\cite{instant4d} also drops the view-time coupling on native 4D Gaussians.
At spatial degree 3 a Gaussian then carries $48 + 24 = 72$ color values instead of $48 + 96 = 144$.
On DyNeRF it matches the 4D harmonics at 45\,\% lower memory (\cref*{S-tab:s_fidelity}); all our models use it.

\paragraph{Densification and optimization in 4D}
We port gradient densification and Selective Adam~\cite{taming3dgs}, both designed for static 3D Gaussians, to 4D.
Ray evaluation has no screen-space positional gradient, so the densification trigger is computed in parameter space as in 3DGRT~\cite{3dgrt}, $\lVert \partial\mathcal{L}/\partial\mu_{xyz} \rVert$ scaled by the distance from the camera to $\mu_{3D}(\tau)$, and averaged only over views whose temporal gate admits the Gaussian, so a short-lived primitive is not diluted.
Selective Adam updates only the Gaussians visible at the rendered time $\tau$, so unlike in 3D, where only the frustum decides, a Gaussian sleeps over time spans, parameters and moments frozen.

\subsection{Beyond the Global-Shutter Pinhole}\label{sec:beyond}

\begin{figure*}[t]
\centering
\includegraphics[width=\textwidth]{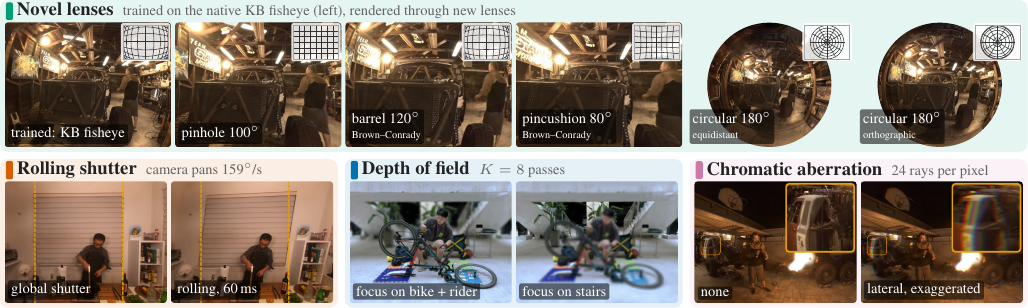}
\caption{Render-time cameras and effects, none of them used in training.
\emph{Novel lenses}: pinhole, barrel, pincushion and two $180^\circ$ circular fisheyes, whose rims lie outside the captured field.
\emph{Rolling shutter}: each row has its own time and pose, so the window frame shears against the dashed global-shutter edges.
\emph{Depth of field}: two focus distances.
\emph{Chromatic aberration}: lateral, exaggerated.}\label{fig:f5}
\end{figure*}

Because every pixel is evaluated as a ray in world space, the camera enters the rasterizer only as a ray and a time per pixel.
Effects that a screen-space splatter can only approximate, per effect and per lens, therefore follow without new primitives or retraining; we show three (\cref{fig:f5}, video on the \href{\projecturl}{project page}; derivations in \cref*{S-sec:s_effects}).

\paragraph{Rolling shutter}
Each image row has its own exposure time $\tau_r$ and pose.
Since $L$ does not depend on time, each Gaussian is conditioned and factored once; a row only moves the mean along its velocity and re-evaluates the temporal gate $w_t$, so row $r$ equals the global-shutter render at $(\tau_r, \mathrm{pose}_r)$ up to fp32 rounding, in one rasterizer pass.

\paragraph{Depth of field}
For a thin lens, the view from an aperture point equals the unchanged camera viewing a laterally sheared scene, so it works through any camera model.
We match the second moment of the aperture's blur disk with a lateral covariance, append its square root to the sheared factor, repeat the QR and rescale the opacity; $K$ such passes at offset aperture points approach exact sampling ($K = 8$ in \cref{fig:f5}).

\paragraph{Chromatic aberration}
A lens bends each wavelength slightly differently, so the color channels see the scene at slightly different magnifications.
Each pixel casts $S$ rays per color channel, one per wavelength, through the lens at that wavelength's magnification, and averages them per channel.
Conditioning, tiles and sort are computed once and shared; a splatter would project every Gaussian once per wavelength.

\section{Experiments}\label{sec:exp}

\begin{figure*}[t]
\centering
\includegraphics[width=\textwidth]{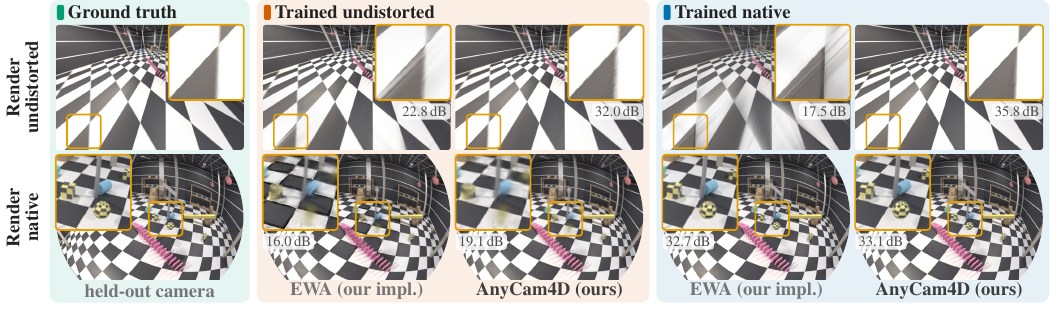}
\caption{Held-out camera on Dynopsis KB; each model is drawn by its own renderer, EWA through the lens Jacobian, and insets enlarge the orange box (PSNR on the crop).
\emph{Trained undistorted:} the 144\,px native center crop covers 48\,px of the undistorted training image, so neither model keeps its detail; at the undistorted rim, $79^\circ$ off axis, EWA smears floor tiles that \modelname keeps sharp.
\emph{Trained native:} both resolve the center; at the undistorted rim the EWA model breaks up, having absorbed EWA's error for the fisheye only.}\label{fig:f4}
\end{figure*}

\subsection{Setup}\label{sec:setup}

\paragraph{Datasets and lenses}
We use four multi-view video datasets that span narrow and wide lenses.
DyNeRF~\cite{li2022neural3dvideosynthesis} (three 300-frame scenes, 20 training cameras, $1352{\times}1014$, camera~0 held out) and the six 60-frame SelfCap clips of FreeTimeGS~\cite{longvolcap,freetimegs} (bike, corgi and dance; 21--23 training cameras and one held out; up to $1920{\times}1080$) are captured with Brown--Conrady (BC)~\cite{brown_distortion,conrady_distortion} lenses, DeepView~\cite{broxton2020immersive} (flames, truck, welder; 150--900 frames; 44 cameras at $1280{\times}960$, one held out) with Kannala--Brandt (KB)~\cite{kannala_brandt} fisheyes.
Dynopsis (\cref*{S-sec:s_dynopsis}), a path-traced scene of 288 frames, is rendered through a BC lens ($800{\times}600$) and a $170^\circ$ KB lens ($800{\times}800$), with 30 training and 2 held-out cameras.
Lenses are quoted by their full field, directions by their incidence from the axis.
We implement pinhole, BC, equidistant and KB (\cref*{S-sec:s_ewa_nat}); any model that maps points to pixels and back can be added.

\paragraph{Baselines}
We reimplement EWA to isolate the effect of projection and rasterization: both renderers, built on gsplat~\cite{gsplat}, share calibration and initial points (\cref*{S-sec:s_init}) and the model, appearance, densification and optimizer of \cref{sec:training}.
Our EWA reaches 34.05\,dB on DyNeRF, against 33.49\,dB reported by 4DGS~\cite{4dgs_yang} (\cref*{S-tab:s_fidelity}).
To separate the renderer from the images it trains on, we also run EWA through the native lens.
This EWA\,nat, an ablation we add and not a published method, projects each mean through the lens and its covariance through the lens Jacobian at the mean (\cref*{S-sec:s_ewa_nat}), as fisheye EWA does for the equidistant lens~\cite{fisheye_gs}; it keeps EWA's depth sort, near-plane cull and screen-space dilation.

\paragraph{Training}
All runs follow the 4DGS schedule~\cite{4dgs_yang}, 30k steps of four renders each with an L1 + 0.2\,D-SSIM loss, with three seeds (details in \cref*{S-sec:s_training}).

\paragraph{Metrics}
We report PSNR, SSIM~\cite{ssim} and LPIPS~\cite{zhang2018lpips} (AlexNet) on the held-out cameras at the last training step, averaged over all frames and seeds, counting only observed pixels (the image circle of a fisheye, the valid region of an undistorted image).
Differences between our models below 0.5\,dB we call ties; seed half-ranges stay below 0.9\,dB apart from the late EWA collapses on flame\_steak (\cref*{S-tab:s_quality}).
Against published methods we follow each dataset's split, frame range, camera model and SSIM variant~\cite{stg}.
Undistorted and native models are scored on different images, so their own scores are not comparable.
We therefore score both on a \emph{shared target}: the part of the native held-out images that the undistorted images also cover, leaving out the rim that only native training sees, with \modelname rendering the undistorted-trained model through the native camera.

\subsection{Comparison with Prior Work}\label{sec:prior}

\Cref{tab:t2} shows that our baseline is sound.
Under each dataset's published protocol, with undistorted images on DyNeRF and SelfCap, our EWA is ahead of 4DGS~\cite{4dgs_yang}, which uses the same primitive, on seven of nine scenes, and \modelname stays close to it.
FreeTimeGS leads on five of the six SelfCap clips it introduced, by 0.2--1.8\,dB.
We build on the 4D Gaussian deliberately: it is the general form, and any FreeTimeGS primitive is a 4D Gaussian in other parameters (\cref*{S-sec:s_freetime}), so our rendering applies to it.
Its lead on fast motion comes from how it finds primitives (4D initialization from matches, relocation, a perceptual loss), which is independent of the projection.
On DeepView, the only benchmark scored on native fisheye images, both renderers lead on truck and welder; published methods train one model per 50-frame segment, favoring them on the long flames clip.

\subsection{Native Training versus Undistortion}\label{sec:native}

At equal pixel count, undistortion trades central resolution against field of view, and Dynopsis KB shows both sides.
Keeping $150^\circ$ across the image lowers the central focal length from 364 to 107\,px, and native training leads by 8.1\,dB on the shared target ($\Delta_\text{nat}$ in \cref{tab:t3}), already at the image center (\cref{fig:f4}, \cref*{S-fig:s_crop}).
Cropping to $95^\circ$ across keeps the native focal length and raises the undistorted model on that field from 23.6 to 29.6\,dB, so resolution explains most of this gap.
The native model still leads there by 3.0\,dB, and the cropped model cannot draw the periphery it was not trained on: over the whole native frame it scores 19.0\,dB, the native model 33.0\,dB (\cref*{S-fig:s_crop}).
Each model's own held-out images hide this trade: there the KB models differ by 0.4\,dB, and on cook\_spinach and Dynopsis BC the sign flips.
EWA\,nat gains almost as much on this lens (7.4\,dB), so the benefit comes from training on native images and not from one renderer.

Milder lenses have less to trade.
The focal length drops by a factor of 1.45 on Dynopsis BC and 1.23 on DeepView, for a native lead of 0.8 and 0.6\,dB; on DyNeRF and SelfCap the difference is $+0.52$ and $-0.15$\,dB.
On DeepView the native model leads inside and beyond the undistorted field alike, while seven further DyNeRF and DeepView scenes are mixed (\cref*{S-sec:s_scenes}), so differences of this size depend on the scene as much as on the lens.
All our rigs are also dense and static: the periphery of one camera is seen near the center of others, which limits what a cropped field can cost.

\subsection{Ray-based Rendering versus EWA}\label{sec:ut-ewa}

\begin{figure}[t]
\centering
\includegraphics[width=\columnwidth]{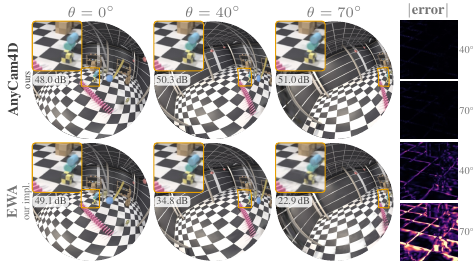}
\caption{Rotation consistency on Dynopsis KB: held-out camera rotated about its center by $\theta$; PSNR and error ($|\Delta|$, black 0 to yellow 0.15) of each native model against its renderer's own unrotated render, which at $\theta = 0$ measure only resampling. Insets resample the orange patch onto one angular grid, so an exact renderer keeps them identical.}\label{fig:rot}
\vspace{8pt}
\includegraphics[width=\columnwidth]{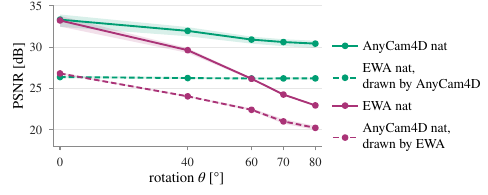}
\caption{PSNR against path-traced ground truth on Dynopsis KB as the held-out camera rotates by $\theta$, on pixels within $40^\circ$ of its unrotated axis; lines: seed mean, bands: range. Path-tracing noise caps PSNR at 31--34\,dB.}\label{fig:rotgt}
\end{figure}

On the same data, the renderers differ where EWA's linearization is stretched.
On the BC lenses they tie in PSNR, apart from late EWA collapses on flame\_steak (2 of 6 EWA runs, none of ours).
On Dynopsis KB, pinhole EWA falls 5.75\,dB behind on undistorted images reaching $79^\circ$ from the axis (\cref{fig:f4}, top) but only 0.57\,dB on the crop reaching $57^\circ$ (\cref*{S-fig:s_crop}), so this deficit belongs to very wide pinhole images.
A Jacobian of the native lens repairs most of it: on the shared target, EWA\,nat trails by 0.68\,dB ($\Delta_\text{proj}$ in \cref{tab:t3}).
On DeepView they tie on flames and truck, and EWA\,nat leads on welder by 0.9\,dB.
Through every native lens, \modelname has lower LPIPS on all 14 scenes, beyond the seed range on 9 (\cref*{S-tab:s_quality}); part may stem from EWA's low-pass filter~\cite{3dgs}.

Part of these ties can come from training rather than from an accurate projection.
A static rig allows this adaptation: every scene point is seen from the same cameras at the same incidences, and the held-out camera sits among them, aimed like its neighbors.
Rotating the held-out camera about its center removes this agreement: every ray stays the same and only moves within the image, so an exact renderer must reproduce its own unrotated render.
Up to a rotation of $80^\circ$, \modelname does so at 29--58\,dB on every model (\cref{fig:rot}), while at $60$--$80^\circ$ EWA falls to 12--35\,dB, on every lens and on undistorted pinhole images alike (\cref*{S-fig:s_rotplot}): the drop follows the change of incidence between training and rendering, not the field of the lens, which only allows larger changes. Two parts of EWA contribute: its screen-space dilation, a fixed blur in pixels and thus a varying one in angle, causes 20--50\,\% of the drop (\cref*{S-sec:s_rotation}), and the linearization of the lens the rest.
Path-traced ground truth (\cref{fig:rotgt}) shows that this is lost accuracy: the native models tie unrotated, and at $60$--$80^\circ$ \modelname leads by 6.2\,dB (5.7--6.5 over seeds).
Drawn by \modelname, the EWA\,nat model stays flat, 6.9\,dB below its unrotated EWA score: the error training absorbed for this rig; drawn by EWA, the {\modelname}\,nat model shows it unabsorbed, 6.5\,dB below already unrotated.
On this scene, an EWA model is accurate only for views aimed like its rig.

\subsection{Cost and Stability}\label{sec:cost}

A training step costs about as much as with EWA.
On the same trained model and views, render plus backward pass of \modelname take 1.06--1.13 times as long as with EWA and the render alone 1.39--1.47 times, since a ray evaluation costs more than a 2D conic (\cref*{S-sec:s_cost}).
Memory per Gaussian is within 3\,\%.
No final run had a non-finite loss or parameter; for \modelname this needs the closest-point offset.

\section{Conclusion}\label{sec:conclusion}

\modelname brings ray-based rendering to native 4D Gaussians: they train on footage as recorded through any camera model, without losing field of view or resolution to undistortion, and render through lenses and camera effects they were not trained with (\cref{fig:f5}, video).
Conditioning the square root of the 4D covariance keeps ray evaluation and its gradients accurate in single precision, at about EWA's cost per training step.
Rotating a held-out camera reveals EWA's error, hidden by the static rig: on the synthetic fisheye, \modelname leads by 6.2\,dB. EWA through the native lens thus suffices near the rig's own views, but not beyond them.

\paragraph{Limitations and future work}
All our rigs are static.
The method does not need one, as a moving camera only adds a pose per frame, and we expect larger gains with moving cameras: they see each point at many incidences, so training cannot absorb EWA's error, and no other camera covers a cropped field.
The camera effects are render-only; moving cameras would also need training through the per-row rolling-shutter render.
Culling at the camera plane limits lenses to $180^\circ$.
Fast motion remains weak: on Dynopsis KB, moving pixels score 17--22\,dB, static ones 30--36\,dB.

\section*{LLM Usage Statement}

The authors used Claude (Anthropic), through Claude Code, as a coding agent for the implementation, to generate experiment scripts, analyses of results, figures, the video and data-preparation code that the authors reviewed and approved, and to support the preparation of this paper and its supplementary material through language and grammar checks and simulated peer reviews. The authors independently evaluated and revised all suggestions and retain full responsibility for the research, its results, and the entire content of this paper.

{
    \small
    \bibliographystyle{ieeenat_fullname}
    \bibliography{main}
}

\clearpage
\twocolumn[{\centering\Large\bfseries Supplementary Material\par\vspace{2em}}]
\setcounter{section}{0}
\setcounter{figure}{0}
\setcounter{table}{0}
\setcounter{equation}{0}
\renewcommand{\thesection}{S\arabic{section}}
\renewcommand{\theequation}{S\arabic{equation}}
\renewcommand{\thetable}{S\arabic{table}}
\renewcommand{\thefigure}{S\arabic{figure}}
\renewcommand{\theHsection}{S\arabic{section}}  %
\renewcommand{\theHequation}{S\arabic{equation}}
\renewcommand{\theHtable}{S\arabic{table}}
\renewcommand{\theHfigure}{S\arabic{figure}}
\section*{Overview}
\Cref{tab:s_overview} lists the sections of this supplement and the parts of the paper they extend.
Section, figure, table and equation numbers with the prefix S refer to this supplement, all others to the paper.

\begin{table}[h]
\centering
\caption{Sections of this supplement and the parts of the paper they extend.}\label{tab:s_overview}
\footnotesize
\setlength{\tabcolsep}{3pt}
\begin{tabular}{@{}lll@{}}
\toprule
& Content & Paper \\
\midrule
\cref{sec:s_setup} & Datasets, training, baselines & \cref*{sec:setup,sec:training} \\
\cref{sec:s_results} & Per-scene results, published methods & Tabs.~\ref*{tab:t3} and \ref*{tab:t2} \\
\cref{sec:s_renderer} & Rotated cameras: consistency, accuracy & \cref*{sec:ut-ewa} \\
\cref{sec:s_factor} & Derivations, numerical measures & \cref*{sec:lift} \\
\cref{sec:s_cost} & Render cost & \cref*{sec:cost} \\
\cref{sec:s_effects} & Rolling shutter, depth of field, & \cref*{sec:beyond} \\
& chromatic aberration, lens morph & \\
\cref{sec:s_gallery} & Pixel budget, unseen lenses, & \cref*{sec:native}, \cref*{fig:teaser} \\
& lens morph and bullet time & \\
\bottomrule
\end{tabular}
\end{table}

\paragraph{Video}
The video on the \href{\projecturl}{project page} starts from the common practice of undistorting real footage to pinhole images.
It shows the resolution that undistortion costs and how elliptical weighted average (EWA) splatting~\cite{ewa,ewa_splatting} through the native lens fails once the camera turns away from the rig.
It then shows how the 3D Gaussian unscented transform (3DGUT)~\cite{3dgut} renders static Gaussians along the ray of each pixel.
This needs a root of each covariance, which the conditioning of 4D Gaussians at a time does not provide.
The factored conditioning of \cref{sec:s_factor} computes this root directly.
With this renderer, \modelname repeats the rotation on Dynopsis and on real footage.
The video then renders trained models through a lens that morphs from a pinhole into a fisheye (\cref{sec:s_morph}), through circular fisheyes of several projections, and with rolling shutter, depth of field and chromatic aberration.
It closes with a summary of quality and cost.

\section{Datasets and Implementation}\label{sec:s_setup}
This section describes the synthetic Dynopsis dataset, the calibration and initial points of every dataset, and the training settings.
It then validates our EWA against the published 4DGS~\cite{4dgs_yang} and derives EWA\,nat, the EWA baseline that renders through the native lens.

\subsection{Dynopsis}\label{sec:s_dynopsis}
We render Dynopsis, a synthetic multi-view video, to test both lens families with an exact calibration and to path-trace ground truth for views outside the rig (\cref{sec:s_rotation_gt}).
It has 288 frames at 24\,fps, path-traced in Blender Cycles with adaptive sampling (at most 128 samples per pixel) and denoising.
The scene is a closed room of $16{\times}16{\times}6$\,m with a 1\,m grid on the walls and ceiling, a checkerboard floor, beams, columns, pipes, shelves and crates, so that straight lines reach the image border in every view.
The moving content covers several speeds and kinds of motion.
Kinematic objects include a pendulum, a rolling sphere, a spinning and a sliding cube, four ceiling fans and a standing fan, all textured so that their rotation is visible.
Rigid-body objects, set in motion by gravity, include a bouncing ball, two colliding cylinders, a toppling column and a cascade of 20 dominoes.

\begin{figure}[t]
\centering
\setlength{\dynw}{\dimexpr\columnwidth*100/828\relax}%
\tikzset{heldout/.style={draw={rgb,255:red,213;green,94;blue,0}, line width=1.2pt}}
\includegraphics{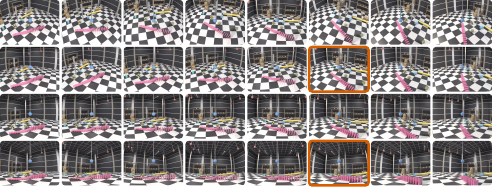}\\[3pt]
\includegraphics{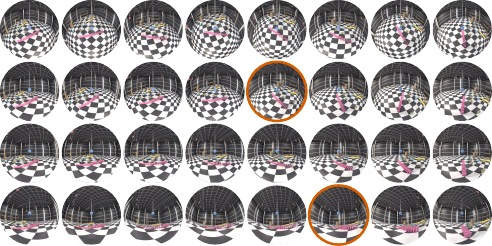}
\caption{Dynopsis at frame 144 through the 32 cameras of the BC rig (top) and the KB rig (bottom), arranged as on the cap with the highest row on top. Orange outlines mark the held-out cameras.}\label{fig:s_dynopsis}
\end{figure}

Each lens has its own rig of 32 cameras on a spherical cap that faces a common target from one side of the room, and two cameras of each rig are held out.
The Brown--Conrady (BC) rig records $800{\times}600$ images with a focal length of 533\,px ($74^\circ$ horizontally before distortion), $k_1 = -0.28$ and $k_2 = 0.06$, from a cap of radius 4\,m that spans $110^\circ{\times}60^\circ$.
The Kannala--Brandt (KB) rig records $800{\times}800$ images with a focal length of 364\,px, $k_1 = -0.25$, $k_2 = 0.06$ and $170^\circ$ across, from a cap of radius 3\,m that spans $140^\circ{\times}80^\circ$.
We path-trace each view once through a camera that covers the whole field of its lens, a pinhole for BC and a $170^\circ$ equidistant fisheye for KB, and resample it to the lens.
As for a real camera, OpenCV undistorts these native images to pinhole images of the same size: for BC to a pinhole that keeps the whole field (focal length 367\,px), for KB to one with $150^\circ$ across the image (107\,px).
\Cref{fig:s_dynopsis} shows every camera of both rigs at one frame.
Dynopsis is available at \url{https://huggingface.co/datasets/sascha-kirch/dynopsis}.

\subsection{Calibration and Initial Points}\label{sec:s_init}
All models of a scene share one calibration and one initial point cloud, whether they train on native or on undistorted images.
For DyNeRF, the published poses assume a distortion-free pinhole, so we run COLMAP~\cite{colmap_sfm} on the first frame of all cameras with one OpenCV camera, a BC model, per video.
COLMAP returns intrinsics, poses and sparse points, which multi-view stereo~\cite{colmap_mvs} densifies to 0.5\,M points.
For DeepView, we keep the published KB calibration; COLMAP triangulates features of the first frame with these poses fixed, and multi-view stereo densifies them to about 1\,M points.
For SelfCap, we use the published calibration and the released dense point cloud closest in time to the clip.
For Dynopsis, we use the exact calibration of the render and back-project the rendered depth of the training cameras.

\subsection{Training}\label{sec:s_training}
\Cref{tab:s_hparams} lists the hyperparameters of our experiments.
As in the published 4DGS code, the temporal learning rate is the spatial one divided by the clip length, and the initial temporal scale is $\sqrt{T/5}$ for a clip of $T$ seconds; we express both in normalized time.

On SelfCap, floaters in front of the held-out camera dominate the error of every method.
We therefore evaluate both renderers with a near plane of 0.3, the value with the best held-out mean in a sweep over 0.1--0.5; FreeTimeGS~\cite{freetimegs} tunes this value per scene.

\begin{table}[t]
\centering
\caption{Training hyperparameters, shared by both renderers and all lenses. The bottom rows give, per dataset, the opacity reset interval, the near plane and how the renders of a step are drawn in time. On DyNeRF, they come from a tile of neighboring frames, since uniform sampling overfits flame\_steak and sear\_steak; on the other datasets, they are drawn uniformly over all frames. LR: learning rate; SH: spherical harmonics; DC: the SH coefficient of degree 0. $^*$Every 10k steps on flame\_steak.}\label{tab:s_hparams}
\footnotesize
\setlength{\tabcolsep}{3pt}
\begin{tabular}{@{}ll@{}}
\toprule
Steps, renders per step & 30k, 4 (truck: 2) \\
Loss & $0.8\,\mathcal{L}_1 + 0.2\,(1 - \text{SSIM})$ \\
Optimizer & Selective Adam \\
Adam $\beta$, $\epsilon$ & $(0.9, 0.999)$, \sci{1}{-15} \\
LR means & $\sci{1.6}{-4} \rightarrow \sci{1.6}{-6}$ (exp.) \\
LR rot.\ / scale / opacity & \sci{1}{-3} / \sci{5}{-3} / \sci{5}{-2} \\
LR color DC / rest & \sci{2.5}{-3} / \sci{1.25}{-4} \\
SH degree, Fourier terms & 3 (+1 every 1k steps), $F = 4$ \\
Initial opacity & 0.1 \\
Densification & steps 500--15k, every 100 \\
Grow / split thresholds & gradient \sci{2}{-4}, scale 0.01 \\
Prune & opacity $< 0.005$, scale $> 0.1$ \\
Temporal prefilter & $w_t < 0.05$ \\
Covariance floor $\varepsilon_c$ & \sci{1}{-6} \\
EWA dilation & 0.3\,px \\
\midrule
DyNeRF & reset 3k$^*$, near 0.01, temporal tiles \\
DeepView, Dynopsis & reset 3k, near 0.01, uniform \\
SelfCap & reset 3k, near 0.1, uniform \\
\bottomrule
\end{tabular}
\end{table}

\subsection{Our EWA Against 4DGS}\label{sec:s_fidelity}
Our EWA baseline reimplements 4DGS.
We validate it against the published 4DGS on DyNeRF (\cref{tab:s_fidelity}), where it reaches a peak signal-to-noise ratio (PSNR) of 34.14\,dB against the published 33.49\,dB.
It departs from 4DGS in two choices of \cref*{sec:training}, which we use for every method we train: additive temporal harmonics instead of multiplicative 4D spherical harmonics, and selective Adam instead of Adam.
Restoring either published choice changes the mean by less than 0.1\,dB, within the seed spread, while the 4D harmonics take 83\,\% more memory.

\begin{table}[t]
\centering
\caption{Our EWA against the published 4DGS on DyNeRF: PSNR on the held-out camera at the last step, mean over three seeds. Each indented row replaces one of our choices by the published one. The first row repeats the EWA of \cref*{tab:t2}, trained on undistorted images, with separate runs, which score 34.05\,dB there. cs, fs, ss: cook\_spinach, flame\_steak, sear\_steak.}\label{tab:s_fidelity}
\footnotesize
\setlength{\tabcolsep}{3pt}
\begin{tabular}{@{}lcccc@{}}
\toprule
& cs & fs & ss & mean \\
\midrule
Our EWA & 34.34 & 33.89 & 34.18 & 34.14 \\
\quad multiplicative 4D harmonics & 34.26 & 33.93 & 34.17 & 34.12 \\
\quad Adam & 34.30 & 34.09 & 34.00 & 34.13 \\
\midrule
4DGS, published & 32.93 & 34.03 & 33.51 & 33.49 \\
\bottomrule
\end{tabular}
\end{table}

\subsection{EWA Through the Native Lens}\label{sec:s_ewa_nat}
EWA\,nat trains EWA on the native images, so that it differs from \modelname trained on the same images only in the projection.
It projects each mean through the lens and its covariance through the Jacobian $J$ of the lens at the mean, $\Sigma_{2D} = J W \Sigma_{3D} W^\top J^\top$, followed by the screen-space dilation of the pinhole path.

Both lens families act on the pinhole point $\bar p = (\bar x, \bar y) = (x/z, y/z)$ of the camera-space mean $(x, y, z)$.
They scale it radially by $\psi(r^2)$, with $r = \lVert \bar p \rVert$, add a tangential term $t(\bar p)$, and map the result to the pixel
\begin{equation}
  m = \operatorname{diag}(f_x, f_y)\,\bigl(\psi(r^2)\,\bar p + t(\bar p)\bigr) + c
\end{equation}
with focal lengths $f_x, f_y$ and principal point $c$.
By the chain rule, $J$ is the product of the Jacobian $J_P$ of the perspective division and the Jacobian $J_D$ of the distortion,
\begin{equation}
\begin{aligned}
  J &= \operatorname{diag}(f_x, f_y)\, J_D\, J_P, \qquad J_P = \frac{1}{z}\begin{pmatrix} 1 & 0 & -\bar x \\ 0 & 1 & -\bar y \end{pmatrix}, \\
  J_D &= \psi\, I + 2\psi'\,\bar p\,\bar p^\top + \partial t/\partial \bar p,
\end{aligned}
\end{equation}
where $\psi' = d\psi/d(r^2)$.
The two lens families differ only in $\psi$ and $t$.
Means beyond the fold of the lens, where the radial map stops increasing, are culled.
Since $J$ depends on the mean, the backward pass also needs the second derivatives of the lens, which are closed forms as well.

For BC with its rational and thin-prism terms, $\psi$ is a ratio of polynomials,
\begin{equation}
\begin{aligned}
  \psi &= \frac{N}{D} = \frac{1 + k_1 r^2 + k_2 r^4 + k_3 r^6}{1 + k_4 r^2 + k_5 r^4 + k_6 r^6}, \\
  t &= \begin{pmatrix} 2p_1 \bar x\bar y + p_2 (r^2 + 2\bar x^2) + s_1 r^2 + s_2 r^4 \\ p_1 (r^2 + 2\bar y^2) + 2p_2 \bar x\bar y + s_3 r^2 + s_4 r^4 \end{pmatrix},
\end{aligned}
\end{equation}
with $\psi' = (N'D - ND')/D^2$.
As in the pinhole path of gsplat~\cite{gsplat}, we evaluate $J$ at $\bar p$ clamped to 1.3 times the half extent of the image in undistorted coordinates, so that means far off screen keep a bounded Jacobian.

For KB, the map is exact for $z > 0$ and has no tangential term:
\begin{equation}
\begin{aligned}
  \psi &= \theta_d(\theta)/r, \qquad \theta = \arctan r, \qquad t = 0, \\
  \theta_d &= \theta\,(1 + k_1\theta^2 + k_2\theta^4 + k_3\theta^6 + k_4\theta^8).
\end{aligned}
\end{equation}
Its derivative is $\psi' = \bigl(\theta_d'/(1+r^2) - \theta_d/r\bigr)/(2r^2)$ with $\theta_d' = d\theta_d/d\theta$; below $r^2 = \sci{1}{-2}$, where this closed form cancels, we use its Taylor series.
KB needs no clamp.
With $k = 0$, the Jacobian reduces to the equidistant one of Fisheye-GS~\cite{fisheye_gs}.

\section{Full Results}\label{sec:s_results}

This section reports the results of every scene, including seven scenes beyond those of the paper, and compares structural and perceptual scores with published methods.

\paragraph{Compared models and targets}
As in the paper, we compare four models of each scene, trained with one of two renderers on one of two kinds of images.
The renderers are \modelname and EWA, the images either the native images of the lens or pinhole images undistorted from them.
EWA\,ud and \modelname\,ud are trained on the undistorted images, EWA\,nat and \modelname\,nat on the native images.
We score the models on the two targets of the paper.
The \emph{undistorted target} compares EWA\,ud and \modelname\,ud on the undistorted held-out images.
The \emph{native target} compares \modelname\,ud, EWA\,nat and \modelname\,nat on the shared target of \cref*{sec:setup}, the part of the native held-out images that the undistorted images also cover; \modelname\,ud is rendered there through the native camera.
All values are means over three seeds at the last step, with $\pm$ half the range over the seeds.
We report PSNR, the structural similarity index SSIM~\cite{ssim} and the learned perceptual image patch similarity LPIPS~\cite{zhang2018lpips} with AlexNet.

\begin{table*}[t]
\centering
\caption{Per-scene quality on the held-out cameras: mean\,$\pm$\,half range over three seeds at the last step. Bold marks scenes on which every seed of \modelname\,nat has a lower LPIPS than every seed of EWA\,nat (\cref*{sec:ut-ewa}). The additional scenes in the lower part use the settings of the main scenes without tuning.}\label{tab:s_quality}
\footnotesize
\setlength{\tabcolsep}{2.2pt}
\begin{tabular}{@{}l ccc ccc cc cc cc@{}}
\toprule
& \multicolumn{6}{c}{Undistorted target} & \multicolumn{6}{c}{Native target (shared)} \\
\cmidrule(lr){2-7}\cmidrule(l){8-13}
& \multicolumn{3}{c}{EWA\,ud} & \multicolumn{3}{c}{\modelname\,ud} & \multicolumn{2}{c}{\modelname\,ud} & \multicolumn{2}{c}{EWA\,nat} & \multicolumn{2}{c}{\modelname\,nat} \\
\cmidrule(lr){2-4}\cmidrule(lr){5-7}\cmidrule(lr){8-9}\cmidrule(lr){10-11}\cmidrule(l){12-13}
Scene & PSNR$\uparrow$ & SSIM$\uparrow$ & LPIPS$\downarrow$ & PSNR$\uparrow$ & SSIM$\uparrow$ & LPIPS$\downarrow$ & PSNR$\uparrow$ & LPIPS$\downarrow$ & PSNR$\uparrow$ & LPIPS$\downarrow$ & PSNR$\uparrow$ & LPIPS$\downarrow$ \\
\midrule
\multicolumn{13}{@{}l}{\emph{DyNeRF}} \\
\quad cook\_spinach & 34.34{\scriptsize\,$\pm$\,0.16} & .968 & .073 & 34.23{\scriptsize\,$\pm$\,0.28} & .968 & .073 & 33.61{\scriptsize\,$\pm$\,0.25} & .088 & 34.02{\scriptsize\,$\pm$\,0.23} & .074 & 34.00{\scriptsize\,$\pm$\,0.16} & .073 \\
\quad flame\_steak & 33.66{\scriptsize\,$\pm$\,0.86} & .972 & .067 & 33.85{\scriptsize\,$\pm$\,0.72} & .971 & .067 & 33.78{\scriptsize\,$\pm$\,0.52} & .081 & 32.03{\scriptsize\,$\pm$\,2.55} & .072 & 34.31{\scriptsize\,$\pm$\,0.26} & .066 \\
\quad sear\_steak & 34.16{\scriptsize\,$\pm$\,0.20} & .972 & .069 & 34.00{\scriptsize\,$\pm$\,0.35} & .973 & .068 & 33.53{\scriptsize\,$\pm$\,0.33} & .081 & 34.06{\scriptsize\,$\pm$\,0.22} & .067 & 34.18{\scriptsize\,$\pm$\,0.44} & .066 \\
\multicolumn{13}{@{}l}{\emph{SelfCap}} \\
\quad bike1 & 25.52{\scriptsize\,$\pm$\,0.14} & .894 & .149 & 25.79{\scriptsize\,$\pm$\,0.48} & .894 & .148 & 25.59{\scriptsize\,$\pm$\,0.46} & .164 & 25.40{\scriptsize\,$\pm$\,0.30} & .156 & 25.26{\scriptsize\,$\pm$\,0.09} & .155 \\
\quad bike2 & 24.24{\scriptsize\,$\pm$\,0.20} & .874 & .165 & 24.05{\scriptsize\,$\pm$\,0.16} & .873 & .165 & 23.90{\scriptsize\,$\pm$\,0.15} & .179 & 23.96{\scriptsize\,$\pm$\,0.15} & .171 & 24.08{\scriptsize\,$\pm$\,0.01} & \textbf{.165} \\
\quad corgi1 & 28.61{\scriptsize\,$\pm$\,0.19} & .871 & .214 & 28.75{\scriptsize\,$\pm$\,0.12} & .872 & .206 & 27.95{\scriptsize\,$\pm$\,0.10} & .247 & 27.42{\scriptsize\,$\pm$\,0.07} & .255 & 27.70{\scriptsize\,$\pm$\,0.10} & \textbf{.238} \\
\quad corgi2 & 28.10{\scriptsize\,$\pm$\,0.16} & .866 & .216 & 28.19{\scriptsize\,$\pm$\,0.05} & .866 & .210 & 27.55{\scriptsize\,$\pm$\,0.05} & .248 & 27.13{\scriptsize\,$\pm$\,0.08} & .254 & 27.27{\scriptsize\,$\pm$\,0.20} & \textbf{.240} \\
\quad dance1 & 27.09{\scriptsize\,$\pm$\,0.24} & .918 & .121 & 27.44{\scriptsize\,$\pm$\,0.17} & .921 & .117 & 26.89{\scriptsize\,$\pm$\,0.15} & .143 & 26.36{\scriptsize\,$\pm$\,0.02} & .136 & 26.65{\scriptsize\,$\pm$\,0.15} & \textbf{.130} \\
\quad dance2 & 26.35{\scriptsize\,$\pm$\,0.06} & .913 & .131 & 26.34{\scriptsize\,$\pm$\,0.26} & .914 & .128 & 25.77{\scriptsize\,$\pm$\,0.23} & .160 & 25.64{\scriptsize\,$\pm$\,0.05} & .148 & 25.78{\scriptsize\,$\pm$\,0.05} & \textbf{.144} \\
\multicolumn{13}{@{}l}{\emph{Dynopsis BC}} \\
\quad scene & 29.72{\scriptsize\,$\pm$\,0.01} & .953 & .034 & 29.90{\scriptsize\,$\pm$\,0.13} & .972 & .032 & 28.63{\scriptsize\,$\pm$\,0.13} & .052 & 29.17{\scriptsize\,$\pm$\,0.02} & .031 & 29.46{\scriptsize\,$\pm$\,0.01} & \textbf{.028} \\
\multicolumn{13}{@{}l}{\emph{DeepView}} \\
\quad flames & 28.14{\scriptsize\,$\pm$\,0.13} & .944 & .096 & 28.77{\scriptsize\,$\pm$\,0.30} & .945 & .093 & 27.41{\scriptsize\,$\pm$\,0.26} & .096 & 27.60{\scriptsize\,$\pm$\,0.25} & .094 & 27.81{\scriptsize\,$\pm$\,0.29} & \textbf{.090} \\
\quad truck & 28.76{\scriptsize\,$\pm$\,0.31} & .935 & .089 & 28.38{\scriptsize\,$\pm$\,0.33} & .937 & .084 & 27.68{\scriptsize\,$\pm$\,0.11} & .085 & 28.58{\scriptsize\,$\pm$\,0.25} & .072 & 28.76{\scriptsize\,$\pm$\,0.17} & \textbf{.067} \\
\quad welder & 27.01{\scriptsize\,$\pm$\,0.80} & .921 & .121 & 26.89{\scriptsize\,$\pm$\,0.29} & .922 & .118 & 26.65{\scriptsize\,$\pm$\,0.32} & .115 & 27.98{\scriptsize\,$\pm$\,0.20} & .095 & 27.08{\scriptsize\,$\pm$\,0.53} & .092 \\
\multicolumn{13}{@{}l}{\emph{Dynopsis KB}} \\
\quad scene & 26.85{\scriptsize\,$\pm$\,0.28} & .949 & .065 & 32.60{\scriptsize\,$\pm$\,0.04} & .975 & .030 & 25.06{\scriptsize\,$\pm$\,0.03} & .046 & 32.43{\scriptsize\,$\pm$\,0.06} & .015 & 33.11{\scriptsize\,$\pm$\,0.22} & \textbf{.012} \\
\multicolumn{13}{@{}l}{\emph{DyNeRF, additional scenes}} \\
\quad coffee\_martini & 27.57{\scriptsize\,$\pm$\,0.14} & .924 & .117 & 27.54{\scriptsize\,$\pm$\,0.41} & .923 & .118 & 27.25{\scriptsize\,$\pm$\,0.39} & .132 & 27.23{\scriptsize\,$\pm$\,0.09} & .117 & 27.01{\scriptsize\,$\pm$\,0.36} & .117 \\
\quad cut\_roasted\_beef & 33.99{\scriptsize\,$\pm$\,0.11} & .968 & .074 & 34.52{\scriptsize\,$\pm$\,0.13} & .969 & .073 & 33.91{\scriptsize\,$\pm$\,0.13} & .087 & 33.99{\scriptsize\,$\pm$\,0.11} & .074 & 34.39{\scriptsize\,$\pm$\,0.19} & .072 \\
\quad flame\_salmon & 28.24{\scriptsize\,$\pm$\,0.26} & .930 & .106 & 28.19{\scriptsize\,$\pm$\,0.07} & .930 & .106 & 27.84{\scriptsize\,$\pm$\,0.07} & .120 & 28.03{\scriptsize\,$\pm$\,0.56} & .106 & 27.59{\scriptsize\,$\pm$\,0.08} & .107 \\
\multicolumn{13}{@{}l}{\emph{DeepView, additional scenes}} \\
\quad exhibit & 20.81{\scriptsize\,$\pm$\,0.51} & .848 & .183 & 21.23{\scriptsize\,$\pm$\,0.53} & .843 & .206 & 22.54{\scriptsize\,$\pm$\,0.44} & .157 & 23.71{\scriptsize\,$\pm$\,0.68} & .139 & 25.35{\scriptsize\,$\pm$\,0.45} & .129 \\
\quad cave & 29.23{\scriptsize\,$\pm$\,1.62} & .924 & .137 & 28.43{\scriptsize\,$\pm$\,0.13} & .920 & .139 & 28.19{\scriptsize\,$\pm$\,0.18} & .130 & 26.96{\scriptsize\,$\pm$\,0.46} & .127 & 24.38{\scriptsize\,$\pm$\,2.89} & .137 \\
\quad face\_paint\_1 & 31.10{\scriptsize\,$\pm$\,0.50} & .964 & .108 & 31.15{\scriptsize\,$\pm$\,0.66} & .966 & .102 & 31.27{\scriptsize\,$\pm$\,0.45} & .080 & 30.68{\scriptsize\,$\pm$\,1.07} & .080 & 30.30{\scriptsize\,$\pm$\,1.00} & \textbf{.077} \\
\quad face\_paint\_2 & 29.62{\scriptsize\,$\pm$\,0.37} & .960 & .089 & 30.27{\scriptsize\,$\pm$\,0.37} & .961 & .087 & 30.23{\scriptsize\,$\pm$\,0.58} & .074 & 30.22{\scriptsize\,$\pm$\,0.34} & .070 & 31.17{\scriptsize\,$\pm$\,0.31} & .069 \\%
\bottomrule
\end{tabular}
\end{table*}

\paragraph{Additional scenes}\label{sec:s_scenes}
The lower part of \cref{tab:s_quality} adds three DyNeRF and four DeepView scenes.
On coffee\_martini and flame\_salmon, our models trail 4DGS by 0.8--1.1\,dB; 4DGS models the view through the windows with an environment map, which we do not implement.
On exhibit, every model collapses at the first opacity reset (peak 25.8--29.4\,dB, last step 20--25\,dB), and tuning did not bring it to the level of published methods.
Its scores therefore reflect the recovery from the reset rather than the projection.
On cave, the seeds differ by up to 2.1\,dB, and on face\_paint\_1, one frame scores about 19\,dB for every model.
On the shared target, native training improves cut\_roasted\_beef, face\_paint\_2 and exhibit, and none of the other additional scenes.

\paragraph{Inside and beyond the undistorted field}
On DeepView, native training also helps beyond the field that the undistorted images cover.
Inside that field, \modelname\,nat leads \modelname\,ud by 0.40, 1.08 and 0.43\,dB on flames, truck and welder, and beyond it by 0.28, 1.76 and 0.15\,dB.
On welder, the lead beyond the field lies within the seed range.

\begin{figure}[t]
\centering
\includegraphics[width=\columnwidth]{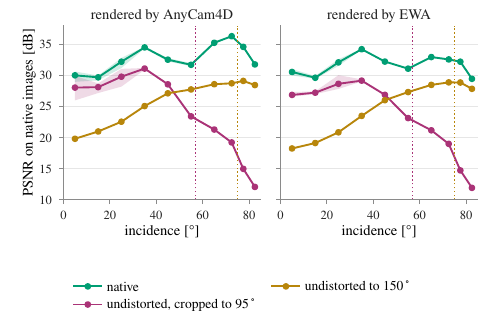}
\caption{PSNR per $10^\circ$ band of incidence over the whole native image of Dynopsis KB, mean and range over three seeds. The models are trained on native images, on undistorted images cropped to $95^\circ$ at the native focal length, and on undistorted images that reach $150^\circ$. Dotted lines mark the corner of the crop ($57^\circ$) and the edge of the $150^\circ$ undistortion ($75^\circ$).}\label{fig:s_crop}
\end{figure}

\paragraph{Resolution against field}
Undistorting Dynopsis KB to $150^\circ$ keeps most of the field but lowers the central focal length from 364 to 107\,px.
\Cref{fig:s_crop} compares it with the opposite choice, an undistortion that keeps the native focal length and with it only $95^\circ$ of the field; \cref{fig:s_budget} shows the three kinds of training images.
On the held-out images of the crop, EWA\,ud trails \modelname\,ud by 0.57\,dB instead of 5.75\,dB.
On the native images, the $150^\circ$ undistortion loses most in the center, where it lowered the resolution.
There, on the central $95^\circ$, the crop recovers 6.0 of the 9.0\,dB that the $150^\circ$ undistortion loses, and the native model still leads by 3.0\,dB.
The crop never sees the rim, so over the whole native image it falls far below both other models (\cref{fig:s_budget}).

\paragraph{SSIM and LPIPS against published methods}
\Cref{tab:s_prior} completes \cref*{tab:t2} with structural and perceptual scores, each in the variant that the reference papers of the dataset use.
On DyNeRF and DeepView, these are the two variants of the structural dissimilarity DSSIM $= (1 - \text{SSIM})/2$ used by Spacetime Gaussians (STGS)~\cite{stg}, together with LPIPS on AlexNet.
This metric is half the term $1 - \text{SSIM}$ of our training loss (\cref{tab:s_hparams}).
DSSIM$_1$ is computed from the Gaussian-window SSIM of 3D Gaussian splatting (3DGS)~\cite{3dgs}, DSSIM$_2$ from the SSIM of scikit-image with a data range of 2, which we call SSIM$_2$.
On SelfCap, we follow FreeTimeGS~\cite{freetimegs} with SSIM$_2$ and LPIPS on VGG.
Our implementations of these metrics match the reference implementations to \sci{1}{-6}.
Our DSSIM is the lowest on DyNeRF and DeepView, consistent with PSNR, while our LPIPS is higher than that of STGS.
LPIPS depends strongly on the evaluation resolution, which the published values do not state: halving the resolution lowers the LPIPS of cook\_spinach from 0.076 to 0.045.
We therefore do not compare LPIPS across papers.

\begin{table*}[t]
\centering
\caption{SSIM-family and LPIPS scores per scene against published methods, in the variant that the reference papers of each dataset use (\cref*{tab:t2} gives PSNR). Published values are taken from the cited papers; ours are seed means at the last step. LPIPS is not comparable across papers (see text). D$_1$, D$_2$: DSSIM$_1$, DSSIM$_2$; S$_2$: SSIM$_2$; LP$_\text{V}$: LPIPS on VGG. $^\dagger$One model per 50-frame segment, trained through a warp of a pinhole render. $^\ddagger$Trained and rendered through the native lens.}\label{tab:s_prior}
\footnotesize
\setlength{\tabcolsep}{3pt}
\begin{tabular}{@{}l ccc ccc ccc@{}}
\toprule
DyNeRF & \multicolumn{3}{c}{cook\_spinach} & \multicolumn{3}{c}{flame\_steak} & \multicolumn{3}{c}{sear\_steak} \\
\cmidrule(lr){2-4}\cmidrule(lr){5-7}\cmidrule(l){8-10}
& D$_1\!\downarrow$ & D$_2\!\downarrow$ & LPIPS$\downarrow$ & D$_1\!\downarrow$ & D$_2\!\downarrow$ & LPIPS$\downarrow$ & D$_1\!\downarrow$ & D$_2\!\downarrow$ & LPIPS$\downarrow$ \\
\midrule
STGS~\cite{stg} & .0215 & .0113 & .037 & .0176 & .0087 & .029 & .0174 & .0085 & .030 \\
Ex4DGS~\cite{ex4dgs} & .0265 & .0120 & .042 & .0220 & .0100 & .034 & .0205 & .0105 & .035 \\
FreeTimeGS~\cite{freetimegs} & .0233 & .0112 & .031 & .0194 & .0087 & .026 & .0198 & .0091 & .026 \\
EWA (our impl.) & .0161 & .0083 & .073 & .0140 & .0070 & .067 & .0141 & .0070 & .069 \\
\modelname{} (ours) & .0158 & .0081 & .073 & .0144 & .0075 & .067 & .0133 & .0065 & .068 \\%
\bottomrule
\end{tabular}

\vspace{0.6em}
\begin{tabular}{@{}l cc cc cc cc cc cc@{}}
\toprule
SelfCap & \multicolumn{2}{c}{bike1} & \multicolumn{2}{c}{bike2} & \multicolumn{2}{c}{corgi1} & \multicolumn{2}{c}{corgi2} & \multicolumn{2}{c}{dance1} & \multicolumn{2}{c}{dance2} \\
\cmidrule(lr){2-3}\cmidrule(lr){4-5}\cmidrule(lr){6-7}\cmidrule(lr){8-9}\cmidrule(lr){10-11}\cmidrule(l){12-13}
& S$_2\!\uparrow$ & LP$_\text{V}\!\downarrow$ & S$_2\!\uparrow$ & LP$_\text{V}\!\downarrow$ & S$_2\!\uparrow$ & LP$_\text{V}\!\downarrow$ & S$_2\!\uparrow$ & LP$_\text{V}\!\downarrow$ & S$_2\!\uparrow$ & LP$_\text{V}\!\downarrow$ & S$_2\!\uparrow$ & LP$_\text{V}\!\downarrow$ \\
\midrule
4DGaussians~\cite{4dgs_wu} & .920 & .312 & .924 & .310 & .900 & .357 & .902 & .365 & .940 & .259 & .936 & .269 \\
STGS~\cite{stg} & .822 & .406 & .827 & .402 & .933 & .219 & .934 & .235 & .933 & .237 & .916 & .245 \\
4DGS~\cite{4dgs_yang} & .919 & .259 & .916 & .269 & .919 & .211 & .924 & .212 & .938 & .244 & .929 & .257 \\
FreeTimeGS~\cite{freetimegs} & .946 & .199 & .937 & .221 & .956 & .158 & .959 & .160 & .953 & .234 & .953 & .232 \\
EWA (our impl.) & .937 & .228 & .919 & .243 & .930 & .230 & .926 & .236 & .947 & .202 & .943 & .206 \\
\modelname{} (ours) & .937 & .227 & .919 & .244 & .930 & .229 & .926 & .237 & .949 & .200 & .943 & .203 \\%
\bottomrule
\end{tabular}

\vspace{0.6em}
\begin{tabular}{@{}l cc cc cc@{}}
\toprule
DeepView & \multicolumn{2}{c}{flames} & \multicolumn{2}{c}{truck} & \multicolumn{2}{c}{welder} \\
\cmidrule(lr){2-3}\cmidrule(lr){4-5}\cmidrule(l){6-7}
& D$_1\!\downarrow$ & LPIPS$\downarrow$ & D$_1\!\downarrow$ & LPIPS$\downarrow$ & D$_1\!\downarrow$ & LPIPS$\downarrow$ \\
\midrule
HyperReel~\cite{hyperreel}$^\dagger$ & .0475 & .159 & .0760 & .223 & .1050 & .281 \\
STGS~\cite{stg}$^\dagger$ & .0349 & .059 & .0524 & .087 & .0504 & .098 \\
EWA (our impl.)$^\ddagger$ & .0325 & .122 & .0387 & .096 & .0395 & .119 \\
\modelname{} (ours)$^\ddagger$ & .0315 & .116 & .0375 & .090 & .0401 & .115 \\%
\bottomrule
\end{tabular}
\end{table*}

\section{Rendering Away from the Training Views}\label{sec:s_renderer}

On the held-out cameras of our rigs, EWA and \modelname mostly tie (\cref*{sec:ut-ewa}).
This section tests whether the tie holds for views that look in other directions than the rig.
EWA linearizes the projection at each Gaussian's mean, so its error depends on where in the image a Gaussian lands, and training adapts the Gaussians until their EWA footprints match the training images.
On a static rig this adaptation transfers to the held-out camera: every scene point is seen by the same cameras at the same incidences, and the held-out camera sits among them, aimed like its neighbors.
A camera that turns, as a handheld or moving camera does, sees the same point at other incidences, where EWA's error differs from the one training compensated.
We isolate this effect by rotating the held-out camera about its center.
\Cref{sec:s_rotation} tests whether each renderer draws a trained model consistently under rotation, and \cref{sec:s_rotation_gt} whether the rotated renders are correct.

\subsection{Consistency Under Rotation}\label{sec:s_rotation}
\paragraph{Protocol}
A rotation about the camera center keeps every ray and moves it to another pixel, and with it to another incidence.
An exact renderer therefore draws each ray the same before and after the rotation, whatever the model.
We rotate the optical axis of the held-out camera by $\theta \in \{5, 10, 20, \dots, 80\}^\circ$ toward four image directions, keeping position, lens, intrinsics and time.
The pixels whose rays lie within $20^\circ$ of the unrotated axis then sit at an incidence of about $\theta$.
We compare each of them with the renderer's unrotated render at twice the resolution, sampled along the same ray.
We pool the squared error over the four directions and five evenly spaced times.
For an exact renderer, the score is the same at every $\theta$, and the score at $5^\circ$ measures the resampling error.
The reference is the renderer's own render, so the score measures how the renderer treats the model, independently of how well the model fits the scene.

\begin{figure}[t]
\centering
\includegraphics[width=\columnwidth]{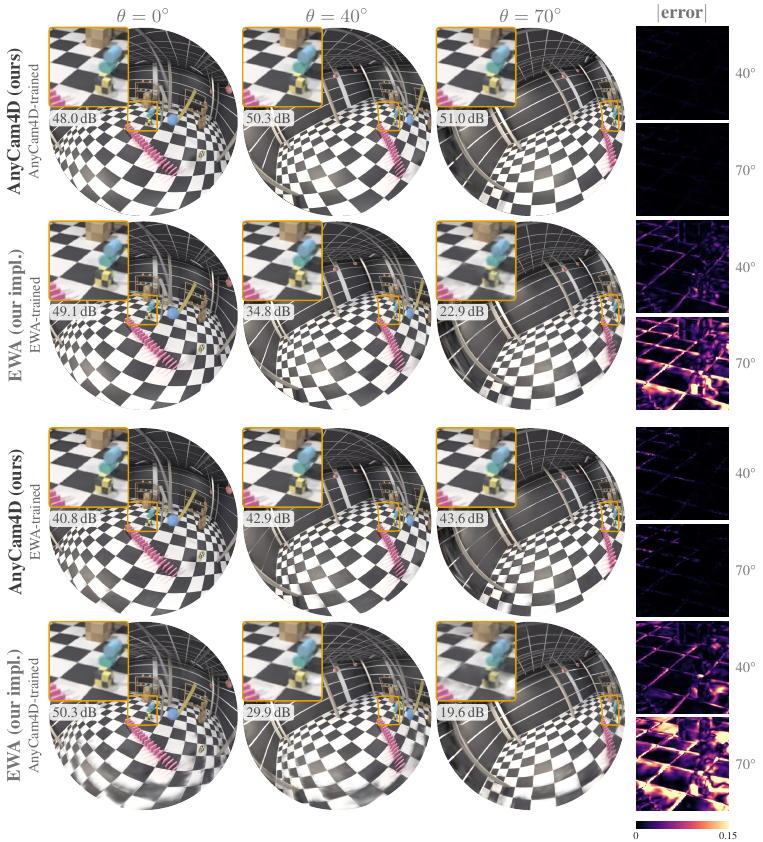}
\caption{Rotation consistency on Dynopsis KB (seed 1). The held-out camera rotates about its center by 0, 40 and $70^\circ$, and the orange outline marks one scene patch, which moves toward the rim. The insets resample the patch onto one fixed angular grid, so an exact renderer gives identical insets; their PSNR is measured against the same renderer's unrotated render along the same rays. Each row shows a model trained with one renderer, drawn by the same or by the other renderer. The right columns show the absolute error at 40 and $70^\circ$.}\label{fig:s_rot}
\end{figure}

\begin{figure*}[t]
\centering
\includegraphics[width=\textwidth]{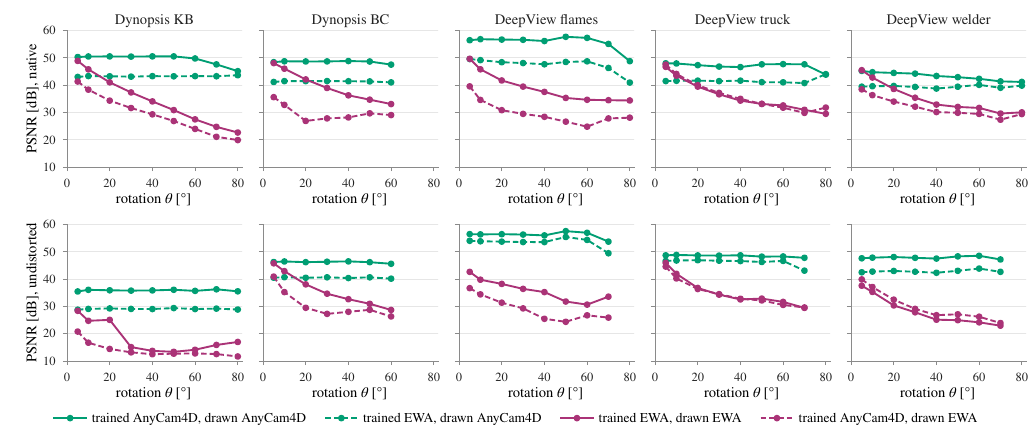}
\caption{Rotation consistency: PSNR of each renderer against its own unrotated render along the same rays, seed mean, for models trained on native (top) and undistorted (bottom) images. Color marks the renderer that draws the rotated view. Solid lines show the renderer the model was trained with, dashed lines the other one. A line ends where no pixel of the reference cone reaches the incidence.}\label{fig:s_rotplot}
\end{figure*}

\paragraph{Results}
\modelname draws every model consistently, including the models trained with EWA (\cref{fig:s_rot,fig:s_rotplot}).
It deviates from its resampling floor only on the native Dynopsis KB model at 70--$80^\circ$, where content approaches the camera plane.
EWA becomes less consistent with every degree of rotation, on every model.
The decline is as clear on the narrow BC lens and on undistorted pinhole images as on the fisheye.
It therefore follows the change of incidence between training and rendering rather than the field of the lens, and a wide lens only allows larger changes.
Training on the static rig bakes EWA's linearization error at the rig's incidences into the Gaussians, and once the camera rotates away from these incidences, the error no longer cancels.
On the models trained with \modelname, which never compensated this error, EWA declines as well, so the error itself changes with incidence.

\subsection{Accuracy Against Path-Traced Rotated Views}\label{sec:s_rotation_gt}
To measure accuracy as well as consistency, we path-trace the rotated views of Dynopsis KB with the renderer of the dataset: both held-out cameras, rotated by $\theta \in \{0, 40, 60, 70, 80\}^\circ$ toward the same four directions, at six frames, 204 images in all.
We score the pixels whose rays lie within $40^\circ$ of the unrotated axis: 161\,k per image unrotated and 67\,k at $80^\circ$, where part of this cone has left the image.
Path-tracing noise limits every score, since the rotated and unrotated ground truth agree along the same rays only at 31--34\,dB.

Unrotated, the native models tie (\cref*{fig:rotgt}), as on the held-out camera of the rig.
Under rotation, \modelname stays close to the limit set by the path-tracing noise, while EWA\,nat falls steadily, 6.2\,dB behind at 60--$80^\circ$ (5.7--6.5 over seeds).
The cross renders locate the error in the model.
Drawn by \modelname, the EWA\,nat model is flat under rotation, which also rules out the shrinking pixel set as the cause, but it starts 6.9\,dB below its own EWA score: its Gaussians are wrong, and only EWA's projection at the incidences of the rig cancels their error.
This measures the adaptation described above: an EWA model is accurate for views aimed like its training cameras and loses accuracy as a view turns away from them.
\modelname\,ud is flat as well, at 22.7--23.6\,dB, the level that the resolution left by undistortion allows.
The experiment covers one scene, one lens and pure rotations of models trained on a static rig; we did not train with a moving camera, which would show each point at more incidences.

\section{Factored Conditioning and Numerics}\label{sec:s_factor}

The paper conditions each 4D Gaussian through a square root of its covariance and factors the result by QR into the triangular factor that the ray evaluation uses (\cref*{sec:lift}).
This section gives the derivations that the paper leaves out and measures their effect on precision in single-precision floating point (fp32), against double precision (fp64) as reference.
\Cref{sec:s_factor_fwd} shows that the factored form is the conditional covariance, also with the temporal floor of the implementation, and derives two bounds of the triangular factor.
\Cref{sec:s_factor_bwd} derives the closed-form gradient back to the 4D Gaussian, and \cref{sec:s_precision} measures the precision of both passes.
\Cref{sec:s_numerics} covers the remaining fp32 measures of the ray evaluation, and \cref{sec:s_freetime} shows that the derivations apply to FreeTimeGS~\cite{freetimegs} as well.

We use the notation of the paper.
The 4D covariance is $\Sigma_{4D} = MM^\top$ with $M = RS$; its three spatial rows form $M_{xyz} \in \mathbb{R}^{3\times4}$ and its temporal row $m \in \mathbb{R}^4$, so that the blocks of \cref*{eq:blocks} are
\begin{equation}\label{eq:s_blocks}
\begin{aligned}
  \Sigma_{xyz} &= M_{xyz}M_{xyz}^\top, \quad \sigma_t^2 = m^\top m, \\
  u &= \Sigma_{xyz,t} = M_{xyz}\,m .
\end{aligned}
\end{equation}
Conditioning on $t = \tau$ (\cref*{eq:cond}) gives $\Sigma_{3D} = \Sigma_{xyz} - uu^\top/\sigma_t^2$.
The paper writes it as $\Sigma_{3D} = NN^\top$ with the $3{\times}4$ root $N$ of \cref*{eq:factored}, through which gradients flow, and factors $N$ by the QR of \cref*{eq:qr} into the lower-triangular $L$ with $LL^\top = NN^\top + \varepsilon_c I$, which the ray evaluation inverts.

\subsection{Forward}\label{sec:s_factor_fwd}

\paragraph{Projector form}
With $\hat m = m/\lVert m\rVert$, the matrix $P = I - \hat m\hat m^\top$ is symmetric and idempotent ($P^2 = P$), hence
\begin{equation}
\begin{aligned}
  M_{xyz} P P^\top M_{xyz}^\top &= M_{xyz}M_{xyz}^\top - \frac{M_{xyz}m\,m^\top M_{xyz}^\top}{m^\top m} \\
  &= \Sigma_{xyz} - \frac{u u^\top}{\sigma_t^2} = \Sigma_{3D},
\end{aligned}
\end{equation}
so $N = M_{xyz}P$ is an exact root of the conditional covariance.

\paragraph{Temporal floor: Potter's update}
The implementation conditions on $\tilde\sigma_t^2 = \sigma_t^2 + \varepsilon_t$ ($\varepsilon_t = \sci{1}{-5}$) in the conditional mean, the conditional covariance and the temporal gate, which bounds the narrowest temporal extent at $\sqrt{\varepsilon_t}$.
The target is then $\Sigma_{xyz} - uu^\top/\tilde\sigma_t^2$, which $P$ no longer produces.
We therefore seek a root of the form $N = M_{xyz}(I - k\,mm^\top)$.
Since $(I - k\,mm^\top)^2 = I - (2k - k^2\sigma_t^2)\,mm^\top$, the condition is $2k - k^2\sigma_t^2 = 1/\tilde\sigma_t^2$, and the root with $k \to 1/\sigma_t^2$ as $\varepsilon_t \to 0$ is
\begin{equation}\label{eq:s_potter}
  k = \frac{1 - \rho}{\sigma_t^2} = \frac{1}{\tilde\sigma_t^2\,(1 + \rho)}, \qquad \rho = \sqrt{\varepsilon_t / \tilde\sigma_t^2} .
\end{equation}
The first form divides by $\sigma_t^2$, which vanishes for $m = 0$, and cancels in $1 - \rho$ for temporally thin Gaussians; the second divides only by $\tilde\sigma_t^2 \ge \varepsilon_t$ and is the one we evaluate.
This is Potter's square-root measurement update~\cite{kaminski_sqrt,bierman}: conditioning on $t = \tau$ is a scalar measurement update, and the temporal floor is its measurement noise.

\paragraph{Triangular factor}
We compute the QR of \cref*{eq:qr} with three Householder reflections and flip the sign of each row of $U$ with a negative diagonal entry, so that $L = U^\top$ is the Cholesky factor of $NN^\top + \varepsilon_c I$.
Householder QR is backward stable: the computed $L$ is the exact factor of a matrix that differs from $A$ by a small multiple of the unit roundoff, relative to $\lVert A\rVert$.
Factoring the Gram matrix $A^\top A$ instead rounds relative to $\lVert A\rVert^2$, which squares the relative error of the thin axis (\cref{tab:s_factor_fwd}).
The floor rows of $A$ give $L$ two bounds.
(i) Every diagonal entry of $L$ is at least $\sqrt{\varepsilon_c}$, so $L$ is always invertible. The entry $L_{kk}$ is the distance of column $k$ of $A$ to the span of the earlier columns, and these vanish in the floor row where column $k$ holds $\sqrt{\varepsilon_c}$.
(ii) $\lVert L^{-1}N\rVert_2 \le 1$, since $L^{-1}(NN^\top + \varepsilon_c I)L^{-\top} = I$.
The backward pass multiplies by $L^{-1}N$, which by (ii) cannot amplify fp32 error.

\subsection{Backward}\label{sec:s_factor_bwd}
The loss reaches the conditioning as $\bar L = \partial\mathcal{L}/\partial L$ from the ray evaluation (\cref*{eq:envelope}), or as $G = \partial\mathcal{L}/\partial\Sigma$ from EWA, which forms $\Sigma = LL^\top$ before projecting it.
We carry it back in two steps, first through the QR to $N$ and then through the conditioning to $M$.

\paragraph{From $L$ to $N$}
$L$ depends on $N$ only through $LL^\top = NN^\top + \varepsilon_c I$, so the adjoint of the QR is that of a Cholesky factorization~\cite{murray_cholesky}.
With $\Phi(X)$ the lower triangle of $X$ with halved diagonal,
\begin{equation}\label{eq:s_nbar}
  \bar N = L^{-\top}\bigl(\Phi + \Phi^\top\bigr)L^{-1}N, \qquad \Phi = \Phi(L^\top \bar L) .
\end{equation}
Evaluated from right to left, it takes two triangular solves per column of $N$ around a $3{\times}3$ product, and its operand $L^{-1}N$ is bounded by (ii).
For EWA, $\bar N = (G + G^\top)N$.

\paragraph{From $N$ to $M$}
With $N = M_{xyz} - k\,u\,m^\top$, $k$ from \cref{eq:s_potter} and $g = \bar N m$, collecting the terms of $\langle \bar N, dN\rangle$ in $dM_{xyz}$ and $dm$ gives
\begin{equation}\label{eq:s_mbar}
\begin{aligned}
  \bar M_{xyz} &= \bar N - k\,g\,m^\top, \\
  \bar m &= -k\,N^\top g - k\,\bar N^\top u + (1 + \rho)\,k^2\,(u^\top g)\,m .
\end{aligned}
\end{equation}
The last term combines the derivative of $k$, $dk/d\sigma_t^2 = -\tfrac12(2 + \rho)\,k^2$, with the term of $d(M_{xyz}m)$, which we rewrite with $M_{xyz}^\top = N^\top + k\,m\,u^\top$.
This keeps the large spatial factor $M_{xyz}$ out of the products, so \cref{eq:s_mbar} loses no more precision than its operands carry.
From $M = RS$ the chain continues to the scales and the two quaternions as for any rotation--scale parameterization.
The conditional mean and the gate involve no subtraction and are differentiated directly.

\subsection{Precision in fp32}\label{sec:s_precision}
The two routes to $L$, through $\Sigma_{3D}$ by \cref*{eq:cond} or through $N$, are equal in exact arithmetic.
In fp32 they differ, but only on Gaussians that are fast and thin, where \cref*{eq:cond} subtracts two nearly equal matrices.
We measure the forward pass on trained models, which contain such Gaussians in realistic numbers, and the backward pass on synthetic ones, where the fp64 gradient is known.

\paragraph{Forward}
For every Gaussian of six checkpoints, one \modelname\,nat model per scene trained with the Schur conditioning, we compare several fp32 routes to the factor with fp64 conditioning and Cholesky factorization.
The measure is the whitened error $\lVert L_{64}^{-1}LL^\top L_{64}^{-\top} - I\rVert_2$, the relative error of every Mahalanobis distance the factor evaluates (\cref{tab:s_factor_fwd}).
Every route through $\Sigma_{3D}$ leaves hundreds to thousands of Gaussians per checkpoint off by more than 1\,\%, some of them by orders of magnitude, whichever factorization follows.
Going through $N$ alone is not enough, since its Gram matrix squares the error back, and only QR on $N$ keeps every Gaussian within a few thousandths.

\begin{table}[t]
\centering
\caption{Forward precision of the conditioned factor: whitened error against fp64 over $N_G$ Gaussians per checkpoint. For the Schur routes, the table gives the number of Gaussians with an error above \sci{1}{-2} and the maximum error; for the factored routes, the maximum error. eigen.: eigendecomposition; Dyn.: Dynopsis; cook\_sp.: cook\_spinach.}\label{tab:s_factor_fwd}
\footnotesize
\setlength{\tabcolsep}{3.4pt}
\begin{tabular}{@{}lr rr rr rr@{}}
\toprule
 & & \multicolumn{4}{c}{Schur $\Sigma_{3D}$} & \multicolumn{2}{c}{factored $N$} \\
\cmidrule(lr){3-6}\cmidrule(l){7-8}
 & & \multicolumn{2}{c}{eigen.} & \multicolumn{2}{c}{Cholesky} & Gram & QR \\
\cmidrule(lr){3-4}\cmidrule(lr){5-6}\cmidrule(lr){7-7}\cmidrule(l){8-8}
Scene & $N_G$ & $>$\sci{1}{-2} & max & $>$\sci{1}{-2} & max & max & max \\
\midrule
flames & 2.50\,M & 5828 & \sci{1}{3} & 3658 & \sci{3}{14} & 1.89 & \sci{4.0}{-3} \\
truck & 2.16\,M & 3110 & 22 & 2828 & 22 & 0.23 & \sci{3.3}{-3} \\
welder & 3.08\,M & 4837 & 18 & 4467 & \sci{1}{8} & 0.38 & \sci{2.4}{-3} \\
Dyn.\ KB & 0.60\,M & 1808 & 5.7 & 1590 & \sci{1}{6} & \sci{5.2}{-2} & \sci{1.8}{-3} \\
Dyn.\ BC & 3.06\,M & 1135 & 8.9 & 1118 & \sci{1}{3} & \sci{1.1}{-2} & \sci{1.3}{-3} \\
cook\_sp. & 1.78\,M & 493 & 3.4 & 474 & 3.4 & \sci{1.2}{-2} & \sci{1.2}{-3} \\
\bottomrule
\end{tabular}
\end{table}

\paragraph{Backward}
We give 20\,000 fast, thin Gaussians (one log-scale in $[1, 3]$, the others in $[-9, -5]$) a random $\bar L$ and compare the per-Gaussian relative gradient error with fp64 automatic differentiation through the Schur form and a Cholesky factorization (\cref{tab:s_factor_bwd}).
For 2837 of them, the fp32 Schur $\Sigma_{3D}$ is not even positive definite.
Differentiating the Schur form cancels as its forward pass does, so its gradient is lost even when its input $\partial\mathcal{L}/\partial\Sigma_{3D}$ is exact.
Going through $N$ removes this loss for either factorization, so the advantage of QR over the Gram route lies in the forward pass.

\begin{table}[t]
\centering
\caption{Backward precision on fast, thin Gaussians: 99th percentile of the per-Gaussian relative gradient error against fp64.}\label{tab:s_factor_bwd}
\footnotesize
\setlength{\tabcolsep}{2.5pt}
\begin{tabular}{@{}lcc@{}}
\toprule
Backward path (fp32) & quats & log-scales \\
\midrule
Schur, from exact fp64 $\partial/\partial\Sigma_{3D}$ & 2.06 & 27.5 \\
Cholesky adjoint, then Schur & 2.03 & 29.8 \\
Through $N$, Gram $NN^\top + \varepsilon_c I$ and Cholesky & \sci{6.2}{-4} & \sci{8.9}{-3} \\
Through $N$, \cref{eq:s_nbar,eq:s_mbar} (ours) & \sci{6.1}{-4} & \sci{7.4}{-3} \\
\bottomrule
\end{tabular}
\end{table}

\paragraph{Effect on training and cost}
The factored conditioning matters in training rather than when rendering a trained model.
Rendered with the factored conditioning, the six checkpoints agree with their Schur renders to at least 72.7\,dB, since the affected Gaussians are too thin to cover many pixels.
In training, however, their gradients reach the whole model.
Of 49 \modelname\,nat trainings with the Schur conditioning, 4 hit non-finite rotation or scale gradients, and none of 72 with the factored one, while the median final PSNR moves by at most $+0.13$\,dB (31 configurations).
The factored conditioning comes at no cost: render and backward take $0.996{\times}$ the time of the Schur conditioning (median of four interleaved processes on two scenes), because the rasterizer no longer factors a covariance per tile.
Peak memory is unchanged.

\subsection{Numerical Measures}\label{sec:s_numerics}
Besides the factored conditioning, the paper takes several measures against fp32 failures (\cref*{sec:lift}).
\Cref{tab:s_numerics} collects all of them with the failure each prevents, measured with and without it.
The paragraphs below add what the paper leaves out about the sigma points, the closest-point offset and the ray clamp.

\begin{table}[t]
\centering
\caption{Numerical measures against fp32 failures and the failure each prevents, measured without and with it. The training counts are small samples on flame\_steak.}\label{tab:s_numerics}
\footnotesize
\setlength{\tabcolsep}{3pt}
\begin{tabular}{@{}>{\raggedright\arraybackslash}p{2.2cm}>{\raggedright\arraybackslash}p{3.1cm}cc@{}}
\toprule
Measure & Failure without it & Without & With \\
\midrule
Triangular factor instead of eigendecomposition & non-finite gradients at isotropic initialization & 37\,\% & 0 \\
\midrule
Closest-point offset & late collapse (training runs) & 4/5 & 0/5 \\
& non-finite runs & 3/15 & 0/10 \\
& relative fp32 error of $D^2$ vs.\ fp64 & 100--2000 & 1 \\
\midrule
Square-root conditioning + QR & Gaussians $>1\,\%$ off in $D^2$ (per checkpoint) & 474--4467 & 0 \\
\midrule
Backward through $N$ & p99 relative error of rotation gradients & 2 & \sci{6}{-4} \\
\midrule
Ray clamp $\ell \ge 0$ & PSNR, undistorted Dynopsis KB & \multicolumn{2}{c}{$+1.8$\,dB} \\
\bottomrule
\end{tabular}
\end{table}

\paragraph{Sigma points from $L$}
Any square root $B$ of $\Sigma_{3D}$ gives sigma points $\mu_{3D} \pm \sqrt{3}\,[B]_i$ with the same mean and covariance, so the unscented transform keeps its second-order accuracy and the choice of factor matters only at higher order.
Against Monte Carlo sampling through an equidistant fisheye up to $100^\circ$ of incidence, the Cholesky factor and the principal axes give the same error of the 2D covariance, about 1\,\% for small Gaussians and a few percent for large ones.

\paragraph{Closest-point offset}
Evaluating a Gaussian maps the offset $p$ from its center to the ray origin into its frame, which divides $p$ by the Gaussian's axes.
For a thin Gaussian far from the camera, the fp32 error of $D^2$ therefore grows with $\operatorname{cond}(L)\,\lVert p \rVert / \sigma_{\min}$, where $\sigma_{\min}$ is the shortest axis.
The camera distance affects every ray-based renderer, 3DGUT~\cite{3dgut} included, while the axis ratio enters because $L$, unlike a rotation, is not orthogonal.
After the move $p \leftarrow p - (p^\top d)\,d$, the offset of every ray the Gaussian can affect is no longer than the Gaussian's own extent, and the error of $D^2$ and of its gradient falls by two to three orders of magnitude for axis ratios of 100--1000.

\paragraph{Ray clamp}
Restricting the evaluation to the ray, $\ell \ge 0$ (\cref*{eq:whiten}), instead of the whole line raises \modelname on undistorted Dynopsis KB by 1.85\,dB.
EWA, which the clamp does not affect, moves by only $+0.05$\,dB between the same two code versions, so the gain is the clamp's.
The clamp matters on wide fields of view. Without it, a Gaussian behind the camera is drawn wherever the line through a pixel passes close to it, and on a wide lens many Gaussians straddle the camera plane.

\subsection{FreeTimeGS Primitives Are 4D Gaussians}\label{sec:s_freetime}
The paper states that our rendering also applies to FreeTimeGS (\cref*{sec:prior}), because a FreeTimeGS primitive, a 3D Gaussian that moves at constant velocity and fades in and out in time, is a 4D Gaussian in other parameters.
We show this here.
At time $t$ it has the mean $\mu_x + v(t-\mu_t)$, a fixed covariance $\Sigma$, and its opacity scaled by $\exp(-\tfrac12 (t-\mu_t)^2/s^2)$.
We choose the blocks of \cref*{eq:blocks} as
\begin{equation}
  \sigma_t^2 = s^2, \qquad \Sigma_{xyz,t} = s^2 v, \qquad \Sigma_{xyz} = \Sigma + s^2 v v^\top .
\end{equation}
This $\Sigma_{4D}$ is positive definite exactly when $\Sigma$ is and $s>0$, since $\Sigma$ is its Schur complement.
\Cref*{eq:cond} then gives $\mu_{3D}(\tau) = \mu_x + v(\tau-\mu_t)$, $\Sigma_{3D} = \Sigma_{xyz} - s^2 v v^\top = \Sigma$ and $w_t = \exp(-\tfrac12(\tau-\mu_t)^2/s^2)$, the same primitive at every time.
Conversely, any 4D Gaussian gives $s^2=\sigma_t^2$, $v=\Sigma_{xyz,t}/\sigma_t^2$ and $\Sigma=\Sigma_{3D}$, so the map is a bijection.
The two methods draw the same family of primitives and differ only in the parameters they optimize, so the factored conditioning, the ray evaluation and the native-lens training apply to FreeTimeGS unchanged.

\section{Render Cost}\label{sec:s_cost}

This section compares the render cost of \modelname and EWA, which supports the statement of \cref*{sec:cost} that a training step of \modelname costs about as much as one with EWA.
We time both renderers on one \modelname\,nat model per real dataset, drawn through the same 40 training views, so that only the renderer differs (\cref{tab:s_cost}).
The renderers share conditioning and spherical harmonics. EWA composites with the tile intersection and rasterizer of gsplat, \modelname with our world-space rasterizer.
We measure on an NVIDIA RTX 4000 Ada Generation Laptop GPU in three processes, each alternating the renderers over six rounds.
Every round starts once the GPU has cooled to $65^\circ$C, every pass starts after four untimed warm-up views, and every view is timed between two synchronizations.
Image transfer, optimizer step and densification are excluded.

Render and backward take 1.06--1.13 times as long with \modelname as with EWA, the render alone 1.39--1.47 times.
The difference lies in the rasterizer.
For every pixel and Gaussian, \modelname finds the closest point on the ray and evaluates the 3D Gaussian there, where EWA evaluates a 2D conic, so its forward rasterization takes about 1.8 times as long.
Its backward pass is about a fifth faster than that of gsplat, which recovers most of the difference in training (per-stage profile of flames on an NVIDIA RTX 2060).
Despite the cool-down, the laptop throttled in about half of the passes, alike for both renderers since they alternate, and excluding these passes moves no ratio by more than 0.08.
The clock of this GPU cannot be locked, so the absolute times hold for this machine only.
Only a small part of each model enters a render, since 18--33\,\% of the Gaussians pass the temporal prefilter at the rendered time and 4--14\,\% are drawn.
\modelname allocates about 55\,MiB more per million Gaussians, about 3\,\% of the memory a Gaussian takes in training.

\begin{table*}[t]
\centering
\caption{Render time per view of one trained model: median over all samples of three processes, and the ratio \modelname/EWA as mean $\pm$ half range over the processes. The Gaussian counts, in millions, give the total, those active at the rendered time and those drawn into the image (medians over the views). Peak is the memory allocated by PyTorch during render with backward.}\label{tab:s_cost}
\footnotesize
\setlength{\tabcolsep}{3pt}
\begin{tabular}{@{}lrrrc rrc rrc rr@{}}
\toprule
& \multicolumn{3}{c}{Gaussians [M]} & & \multicolumn{3}{c}{render [ms]} & \multicolumn{3}{c}{render + L1/SSIM + backward [ms]} & \multicolumn{2}{c}{peak [GiB]} \\
\cmidrule(lr){2-4}\cmidrule(lr){6-8}\cmidrule(lr){9-11}\cmidrule(l){12-13}
Scene & total & active & drawn & Pixels & \modelname & EWA & ratio & \modelname & EWA & ratio & \modelname & EWA \\
\midrule
flame\_steak & 2.18 & 0.55 & 0.31 & 1352${\times}$1014 & 6.2 & 4.5 & 1.39\,$\pm$\,0.02 & 18.9 & 17.9 & 1.06\,$\pm$\,0.02 & 2.31 & 2.19 \\
flames & 2.95 & 0.54 & 0.12 & 1280${\times}$960 & 5.4 & 3.8 & 1.43\,$\pm$\,0.04 & 15.4 & 13.7 & 1.13\,$\pm$\,0.01 & 2.80 & 2.64 \\
corgi1 & 7.05 & 2.32 & 0.51 & 1920${\times}$1080 & 14.4 & 9.8 & 1.47\,$\pm$\,0.02 & 41.0 & 36.4 & 1.12\,$\pm$\,0.01 & 6.31 & 5.93 \\%
\bottomrule
\end{tabular}
\end{table*}

\section{Camera Effects}\label{sec:s_effects}
\newsavebox{\efxbox}
\newcommand{\efx}[3]{\sbox{\efxbox}{\includegraphics[width=#1]{supp/figs/effects/#2}}\begin{tikzpicture}
\begin{scope}\clip[rounded corners=1.5pt] (0,0) rectangle (\wd\efxbox,\ht\efxbox);\node[anchor=south west, inner sep=0] at (0,0) {\usebox{\efxbox}};\end{scope}
\node[anchor=south west, fill=black, fill opacity=0.55, text opacity=1, text=white, font=\scriptsize, inner sep=2pt, rounded corners=1.5pt] at (2pt,2pt) {#3};\end{tikzpicture}}
The paper renders rolling shutter, depth of field and chromatic aberration through trained models (\cref*{sec:beyond}, \cref*{fig:f5}).
This section derives how each effect is computed and where it approximates, and describes the lens morph of the video on the project page.
Each effect only changes the rays and times handed to the rasterizer, so it works through every camera model.
None of them is used in training.

\subsection{Rolling Shutter}
\begin{figure}[t]
\centering
\newcommand{\rsrow}[1]{\efx{0.327\columnwidth}{rs_#1_global.jpg}{global}\hfill\efx{0.327\columnwidth}{rs_#1_pan.jpg}{pan}\hfill\efx{0.327\columnwidth}{rs_#1_jello.jpg}{shake}}
\includegraphics{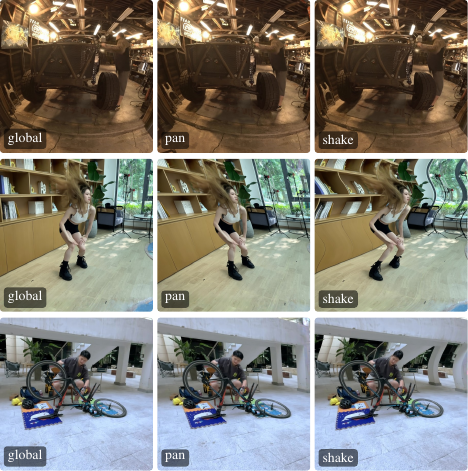}
\caption{Rolling shutter with a moving camera and a real readout of 33\,ms: truck through its KB fisheye, dance1 and bike1, each through a training camera with its focal length shortened by a fifth, at a training instant. The left column shows a global shutter. In the middle column, a pan at $450^\circ$/s slants vertical lines; in the right column, a shake of $\pm1.5^\circ$ at 90\,Hz, three periods per readout, makes straight lines wave.}\label{fig:s_rs}
\end{figure}

With a readout time $\tau_{\mathrm{ro}}$, row $r$ of an $H$-row image is read out at $\tau_r = \tau + \bigl((r + \tfrac12)/H - \tfrac12\bigr)\,\tau_{\mathrm{ro}}$, so that the center row has the frame time, and each row has its own pose.
The conditional mean is linear in time and $L$ does not depend on it, so one conditioning at $\tau$, with the velocity $v = \Sigma_{xyz,t}/\sigma_t^2$ (\cref*{eq:cond}), gives every row:
\begin{equation}
\begin{aligned}
  \mu_{3D}(\tau_r) &= \mu_{3D}(\tau) + v\,(\tau_r - \tau), \\
  w_t(\tau_r) &= \exp\bigl(-(\tau_r - \mu_t)^2 / 2\sigma_t^2\bigr) .
\end{aligned}
\end{equation}
The projection repeats the unscented projection for every row in which the gate admits the Gaussian, with the sigma points moved and seen from that row's pose, and assigns the Gaussian the tiles of all these rows; the rasterizer casts each ray from the pose of its row.
Row $r$ thus equals the global-shutter render at $(\tau_r, \mathrm{pose}_r)$ up to fp32 rounding, which we confirmed against a per-row reference.
Unlike the rolling shutter of 3DGUT~\cite{3dgut} for static scenes, this needs no iteration per point.
Two approximations remain.
Each Gaussian keeps one depth key, which changes the color only where the order flips, by at most 0.0025 at readouts five times longer than real ones, and its color is evaluated once, at $\tau$.
In \cref{fig:s_rs} the camera turns during the readout, so every row sees the scene from its own pose and instant.
The rows fall between the training frames, so the image inherits the temporal interpolation of the model, and short-lived primitives cross-fade.

\subsection{Depth of Field}
In camera coordinates, the aperture is a disk of radius $A$ in the plane $z = 0$ and the focus plane is $z = \delta$.
The ray from an aperture point $o$ through a point $p$ meets the focus plane where the center pinhole sees the sheared point $p' = \bigl(p_{xy} - o_{xy}(1 - z/\delta),\, z\bigr)$, so one aperture sample is the unchanged camera viewing a sheared scene.
For a pinhole, this is the accumulation-buffer depth of field~\cite{haeberli1990accumulation}. For any other lens, we take the shear as the definition, which is exact wherever the lens is locally a pinhole.
The shear is linear, so a Gaussian with camera-space covariance $\Sigma_c$ shears exactly to $Z_o\Sigma_c Z_o^\top$, where $Z_o$ is the shear of the aperture point $o$.

\begin{figure}[t]
\centering
\includegraphics{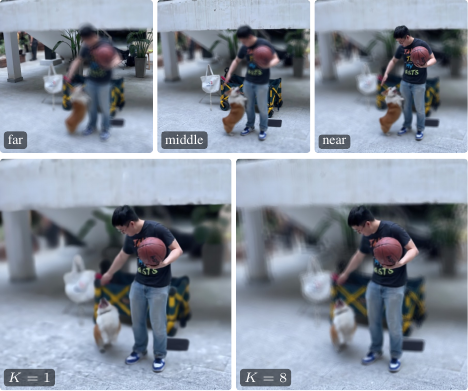}
\caption{Depth of field on SelfCap corgi1 through a KB lens halfway between a pinhole and an equidistant fisheye, which the model was not trained with. The top row pulls the focus from the background to the person ($K = 6$ passes); the bottom row focuses on the person with one and with eight passes.}\label{fig:s_dof}
\end{figure}

Averaged over the aperture, the shear spreads each point over a disk of radius $r(z) = A\,\lvert 1 - z/\delta \rvert$ and covariance $\tfrac{r^2}{4} I_2$.
Matching this second moment at the Gaussian's mean depth $\bar z$ dilates its covariance,
\begin{equation}
\begin{aligned}
  \Sigma_c' &= \Sigma_c + \frac{r(\bar z)^2}{4\kappa}\,\mathrm{diag}(1, 1, 0), \\
  \kappa &= 1 - \frac{(-\ln\rho)\,\rho}{1 - \rho} \approx 0.949,
\end{aligned}
\end{equation}
where $\kappa$ restores the variance that a 2D Gaussian loses when the rasterizer cuts it off at the relative density $\rho = e^{-9/2}$.
We append the square root of the dilation to the factor and repeat the QR, as for the floor $\varepsilon_c$, and scale the opacity by $\bigl(d^\top \operatorname{adj}(\Sigma)\, d \,/\, d^\top \operatorname{adj}(\Sigma')\, d\bigr)^{1/2}$, so that the footprint along each ray direction $d$ keeps its integral.

One blurred pass composites blurred layers, whereas the true image averages sharp composites, and the two differ where blurred layers overlap with contrast.
We therefore render $K$ passes at aperture offsets of covariance $\tfrac{A^2}{4}(1 - \tfrac1K) I$, each blurred with radius $A/\sqrt K$. They keep the second moment of the aperture and move from the single pass ($K = 1$) to aperture sampling ($K \to \infty$).
Against averaged aperture samples (\cref{tab:s_dof}), a single pass loses accuracy as the blur grows and on the fisheye, where more blurred layers overlap, and eight passes gain 7--15\,dB.
The remaining gap on Dynopsis KB comes from the compositing order, since a pinhole camera at the same pose shows the same gap.
In \cref{fig:s_dof}, the single pass draws the background crisper than eight passes, which lie closer to the aperture average.
Three approximations remain. Each Gaussian has one blur radius, and the blur is a Gaussian rather than the shape of the aperture. Beyond about $60^\circ$ on a wide fisheye, where the aperture is seen nearly edge-on, the radial blur is overestimated, by $7\,\%$ at $60^\circ$.

\begin{table}[t]
\centering
\caption{Depth of field with $K$ passes against the average of 256 exact aperture samples: PSNR in dB for a blur radius at infinity of 5\,/\,10\,/\,20\,px.}\label{tab:s_dof}
\footnotesize
\begin{tabular}{@{}lcc@{}}
\toprule
Scene, lens & $K = 1$ & $K = 8$ \\
\midrule
flame\_steak, pinhole & 55.0\,/\,47.7\,/\,42.4 & 70.0\,/\,61.7\,/\,54.9 \\
Dynopsis KB & 32.4\,/\,27.0\,/\,24.3 & 45.1\,/\,38.7\,/\,31.6 \\
\bottomrule
\end{tabular}
\end{table}

\subsection{Chromatic Aberration}\label{sec:s_ca}
\begin{figure}[t]
\centering
\includegraphics{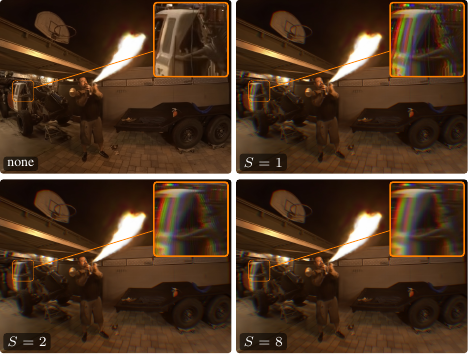}
\caption{Lateral chromatic aberration on DeepView flames through its native KB lens, at the strength of \cref*{fig:f5} ($s = 2.4\,\%$), with $S$ wavelengths per color channel.}\label{fig:s_ca}
\end{figure}

We model lateral chromatic aberration by scaling the focal length about the principal point with a Cauchy dispersion curve, $m(\lambda) = 1 + s\,(\lambda_G^{-2} - \lambda^{-2})/(\lambda_B^{-2} - \lambda_R^{-2})$, so that $m = 1$ at green ($\lambda_G = 535$\,nm) and the magnifications of red and blue ($\lambda_R = 630$\,nm, $\lambda_B = 455$\,nm) differ by $s$.
Each pixel casts $S$ rays per color channel, at wavelengths stratified across the band of the channel, and averages them. All $3S$ rays share conditioning, color, depth order and tiles.
The RGB radiance of the model stands in for the spectrum within each band.
With one wavelength per channel, an edge splits into three shifted copies, and more wavelengths spread it into the continuous fringe of a real lens (\cref{fig:s_ca}).

\newsavebox{\cabox}\newlength{\caw}\newlength{\cah}
\newcommand{\cazoom}[2]{\sbox{\cabox}{\includegraphics[width=0.495\columnwidth]{supp/gallery/ca/#1}}%
\setlength{\caw}{\wd\cabox}\setlength{\cah}{\ht\cabox}%
\begin{tikzpicture}[spy using outlines={rectangle, rounded corners=1.5pt, magnification=3.5, width=0.32\caw, height=0.32\caw, draw=orange, line width=0.6pt, connect spies}]
\begin{scope}\clip[rounded corners=1.5pt] (0,0) rectangle (\caw,\cah);\node[anchor=south west, inner sep=0] at (0,0) {\usebox{\cabox}};\end{scope}
\spy on (0.100\caw,0.475\cah) in node[anchor=north west] at (0.665\caw,0.985\cah);
\node[anchor=south west, fill=black, fill opacity=0.55, text opacity=1, text=white, font=\scriptsize, inner sep=2pt, rounded corners=1.5pt] at (0.015\caw,0.015\cah) {#2};
\end{tikzpicture}}

\newcommand{\zt}[2]{\begin{minipage}[t]{0.327\columnwidth}\centering
\begin{tikzpicture}
\sbox{\efxbox}{\includegraphics[width=\linewidth]{supp/gallery/plates/zoo_#1}}%
\begin{scope}\clip[rounded corners=1.5pt] (0,0) rectangle (\wd\efxbox,\ht\efxbox);\node[anchor=south west, inner sep=0] at (0,0) {\usebox{\efxbox}};\end{scope}
\begin{scope}[shift={(0.03\linewidth,0.03\linewidth)}, x=0.26\linewidth, y=0.26\linewidth]
\fill[white, fill opacity=0.85, rounded corners=1pt] (-0.04,-0.04) rectangle (1.04,1.04);
\begin{scope}\clip (0,0) rectangle (1,1);
\begin{scope}[line width=0.35pt, draw=black!75]\input{supp/gallery/plates/grid_#1}\end{scope}\end{scope}
\draw[black!40, line width=0.3pt] (0,0) rectangle (1,1);
\end{scope}
\end{tikzpicture}\\[-1pt]{\scriptsize #2}\end{minipage}}

\begin{figure*}[t]
\begin{minipage}[t]{\columnwidth}\vspace{0pt}
\centering
\includegraphics[width=0.95\columnwidth]{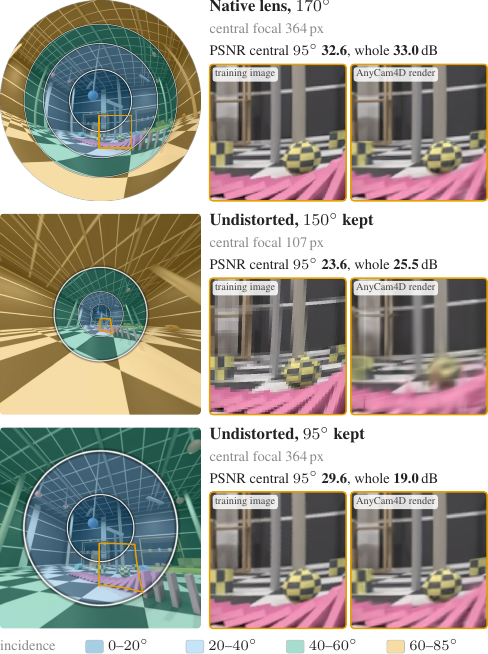}
\caption{The pixel budget of a $170^\circ$ fisheye: held-out camera of Dynopsis KB, $800{\times}800$\,px. The left column shows each kind of training image, tinted by the angle of incidence. Undistorting to an $800{\times}800$ pinhole that keeps $150^\circ$ lowers the central focal length from 364 to 107\,px, and keeping the native focal length leaves $95^\circ$. The right columns show the orange box at the pixels of each image and the render of \modelname, trained on these images, through the native lens. PSNR on the native images, mean of three seeds.}\label{fig:s_budget}
\end{minipage}\hfill
\begin{minipage}[t]{\columnwidth}\vspace{0pt}
\centering
\includegraphics{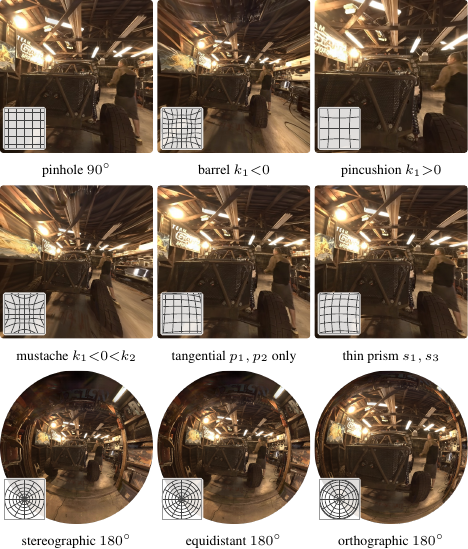}
\caption{One trained \modelname model rendered through nine lens models not used in training, from the same pose and time. The top two rows show OpenCV pinhole-family models, the bottom row the classic fisheye mappings $r = 2f\tan\frac{\theta}{2}$, $f\theta$ and $f\sin\theta$ as KB coefficients fitted to within \sci{1}{-4} of $r/f$. Each inset shows the image of a straight-line grid through the lens, and for the circular fisheyes the image of rings every $15^\circ$ of incidence and spokes every $30^\circ$.}\label{fig:s_zoo}
\end{minipage}
\end{figure*}

\subsection{Lens Morph}\label{sec:s_morph}
The video on the project page and \cref{fig:s_strip_morph} morph one lens continuously from a pinhole into a fisheye while the scene plays.
We describe the lens with a single KB model and blend its coefficients.
Since $\tan\theta = \theta\,(1 + \tfrac13\theta^2 + \tfrac{2}{15}\theta^4 + \tfrac{17}{315}\theta^6 + \tfrac{62}{2835}\theta^8 + \dots)$, the coefficients $k_\text{tan} = (\tfrac13, \tfrac{2}{15}, \tfrac{17}{315}, \tfrac{62}{2835})$ make KB a pinhole up to the truncated series (radial error 0.1\,\% at $45^\circ$).
The series of $\sin\theta$ gives $k_\text{sin} = (-\tfrac16, \tfrac{1}{120}, -\tfrac{1}{5040}, \tfrac{1}{362880})$ for the orthographic fisheye $r = f\sin\theta$, and $k = 0$ is equidistant.
At morph state $\gamma \in [0,1]$, we use $k = (1-\gamma)\,k_\text{tan} + \gamma\,k_\text{end}$ and set the focal length so that the image diagonal spans a field interpolated linearly between the start and the end lens.
Every intermediate camera is a valid KB lens, so \modelname renders all of them with the trained model, without retraining.
Fixed lenses that the model was not trained with work the same way, and \cref{fig:s_zoo} in the gallery shows one model through nine of them.

\section{Gallery}\label{sec:s_gallery}
\begin{figure*}[p]
\centering
\includegraphics{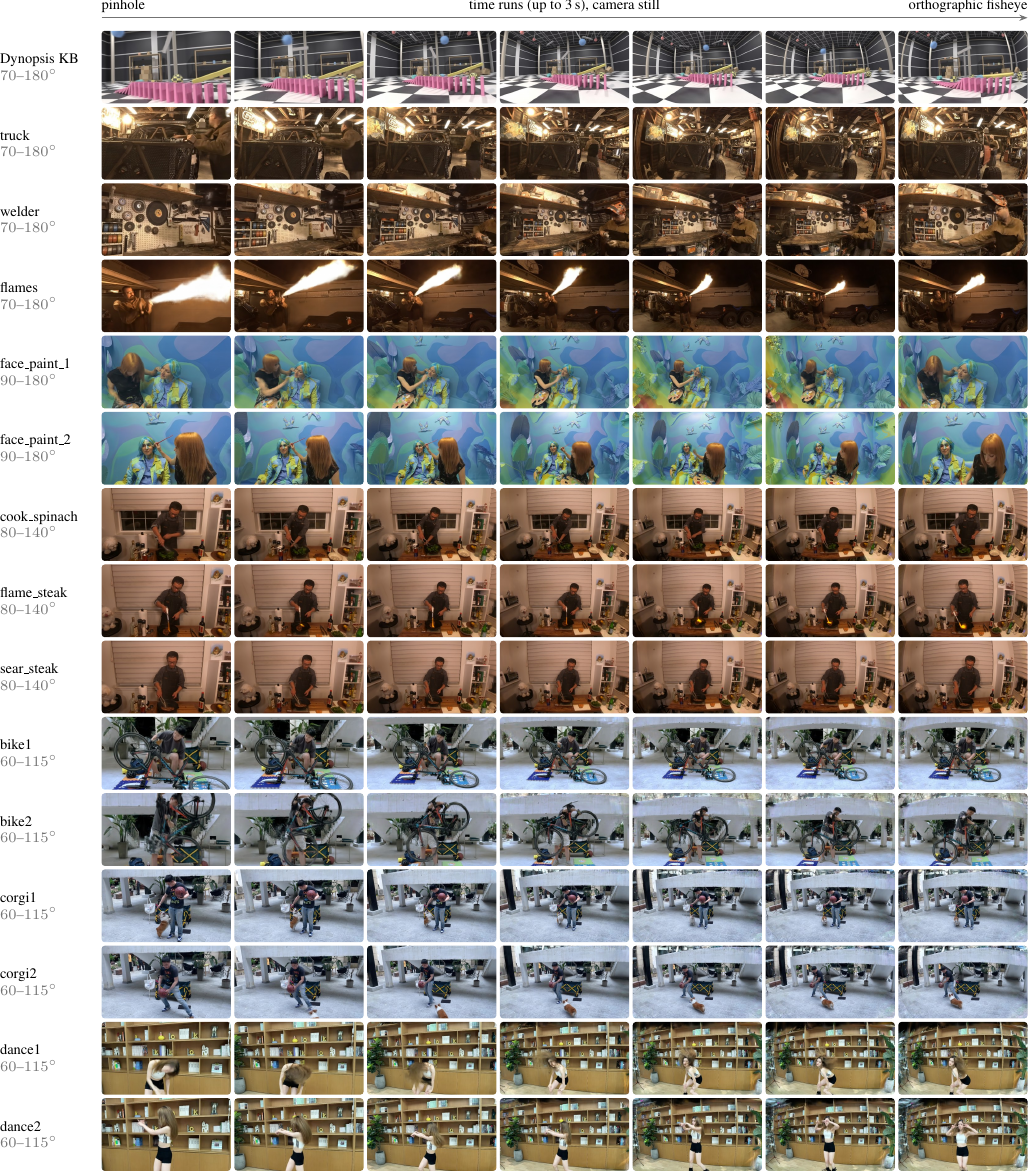}
\caption{Lens morph on every scene. Each row is one trained \modelname model, rendered left to right from a fixed camera through a lens that opens from a pinhole to an orthographic fisheye, $r = f\sin\theta$, while time runs. The final field (range at left) is as wide as the rig's cameras together cover. Every image is a single render through that lens. Truck and welder are models trained on the short clips of the video on the project page.}\label{fig:s_strip_morph}
\end{figure*}

\begin{figure*}[p]
\centering
\includegraphics{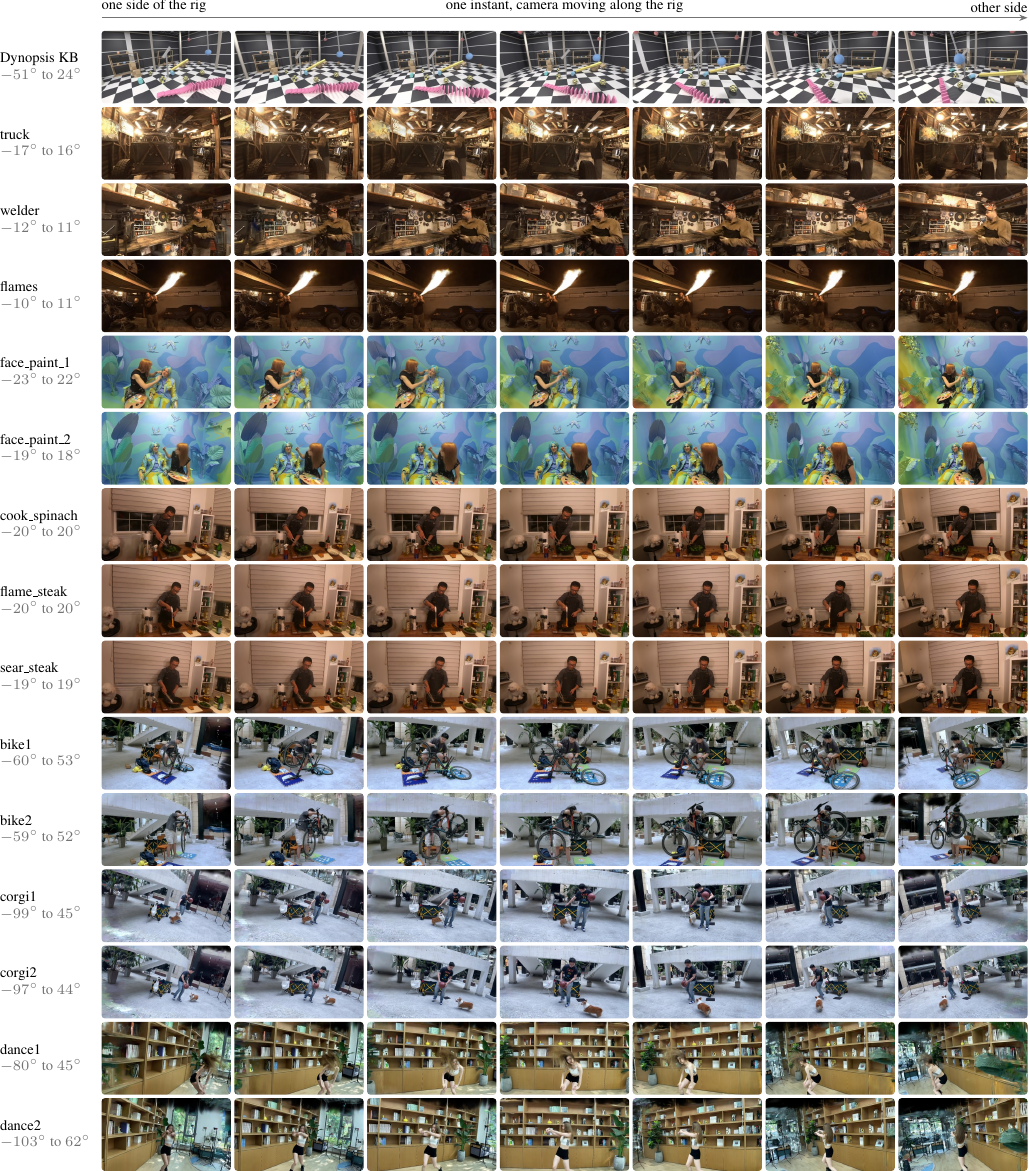}
\caption{Bullet time on every scene. Each row freezes one instant in the middle of the clip and moves the camera along a smooth path through the rig's training cameras, from the second camera on one side to the second on the other, always looking at the subject; the range at left is the path's yaw about the subject. One lens per rig: $120^\circ$ diagonal for DeepView and Dynopsis, $95^\circ$ for DyNeRF, $85^\circ$ for SelfCap.}\label{fig:s_strip_bullet}
\end{figure*}

This section shows renders of trained models beyond the evaluation.
\Cref{fig:s_budget} shows why we train on the native lens: at equal pixel count, a pinhole image of a $170^\circ$ fisheye gives up either resolution at the center or field at the rim.
\Cref{fig:s_zoo} renders one trained model through nine lens models not used in training, from the same pose and time.
\Cref{fig:s_strip_morph,fig:s_strip_bullet} apply the lens morph of \cref{sec:s_morph} and a bullet-time camera path to every scene.

\newcount\striprowi
\def\SL{1.72}\def\SG{0.06}
\pgfmathsetmacro{\SW}{(17.4-\SL-6*\SG)/7}
\pgfmathsetmacro{\SH}{\SW*9/16}
\newcommand{\striprow}[4]{%
  \pgfmathsetmacro{\sy}{-\the\striprowi*(\SH+\SG)}
  \node[anchor=west, font=\scriptsize, align=left, inner sep=0pt] at (0,{\sy-\SH/2}) {#3\\{\color{black!55}#4}};
  \foreach \c in {0,...,6}{%
    \pgfmathsetmacro{\sx}{\SL+\c*(\SW+\SG)}
    \begin{scope}\clip[rounded corners=1.5pt] (\sx,\sy) rectangle ++(\SW,-\SH);
    \node[anchor=north west, inner sep=0pt] at (\sx,\sy) {\includegraphics[width=\SW cm]{supp/figs/strips/#1_#2_\c}};\end{scope}}
  \global\advance\striprowi by 1}
\newcommand{\striphead}[3]{%
  \draw[-{Stealth[length=4pt]}, black!55, line width=0.5pt] (\SL,0.22) -- ({\SL+7*\SW+6*\SG},0.22);
  \node[anchor=south west, font=\scriptsize, inner sep=0pt] at (\SL,0.3) {#1};
  \node[anchor=south, font=\scriptsize, inner sep=0pt] at ({\SL+3.5*\SW+3*\SG},0.3) {#2};
  \node[anchor=south east, font=\scriptsize, inner sep=0pt] at ({\SL+7*\SW+6*\SG},0.3) {#3};}

\end{document}